\documentclass[letterpaper]{article} 
\usepackage{aaai23}  
\usepackage{times}  
\usepackage{helvet}  
\usepackage{courier}  
\usepackage[hyphens]{url}  
\usepackage{graphicx} 
\usepackage{natbib}  
\usepackage{caption} 
\usepackage{algorithm}
\usepackage{algpseudocode}
\usepackage{enumerate}
\usepackage{amssymb}
\usepackage{amsfonts}
\usepackage{amsmath}
\usepackage{mathtools}
\usepackage{xcolor}
\usepackage{multicol}
\usepackage{multirow}
\usepackage{booktabs}
\usepackage{tikz}
\usepackage{pgfplots}
\usepackage{enumitem}
\usepackage{subfig}

\usepackage[linecolor=orange,size=scriptsize]{todonotes}

\newcommand\Tstrut{\rule{0pt}{2.6ex}}       
\newcommand\Bstrut{\rule[-0.9ex]{0pt}{0pt}} 
\newcommand{\TBstrut}{\Tstrut\Bstrut} 

\definecolor{darkgray2}{rgb}{0.36, 0.36, 0.36}
\definecolor{LightCyan}{rgb}{0.8,0.9,0.8}
\definecolor{LightRed}{rgb}{1,0.75,0.75}
\definecolor{teal}{rgb}{0.98, 0.75, 0}
\definecolor{Gray}{gray}{0.93}
\definecolor{mintbg}{rgb}{.63,.79,.95}
\definecolor{gblue}{RGB}{29, 144, 255}

\newcommand{\light}[1]{\textcolor{darkgray2}{#1}}

\usepackage{newfloat}
\usepackage{listings}
\DeclareCaptionStyle{ruled}{labelfont=normalfont,labelsep=colon,strut=off} 
\floatstyle{ruled}
\newfloat{listing}{tb}{lst}{}
\floatname{listing}{Listing}
\newcommand{\X}{{FaIRL}}

\title{Sustaining Fairness via Incremental  Learning}
\author{
    Somnath Basu Roy Chowdhury \qquad
    Snigdha Chaturvedi\\
}
\affiliations{
    \texttt{\{somnath, snigdha\}@cs.unc.edu}\\
    UNC Chapel Hill
    
}

\usepackage{bibentry}

\begin{document}

\maketitle

\begin{abstract}
Machine learning systems are often deployed for making critical decisions like credit lending, hiring, etc. While making decisions, such systems can encode the user's demographic information (like gender, age) in their intermediate representations. This can lead to decisions that are biased towards specific demographics. Prior work has focused on  debiasing intermediate representations to ensure fair decisions. However, these approaches fail to remain fair with changes in the task or demographic distribution. To ensure fairness in the wild, it is important for a system to adapt to such changes as it accesses new tasks in an incremental fashion.  In this work, we propose to address this issue by introducing the problem of learning fair representations in an incremental learning setting. To this end, we present \textbf{F}airness-\textbf{a}ware \textbf{I}ncremental \textbf{R}epresentation \textbf{L}earning ({\X}), a representation learning system that can sustain fairness while incrementally learning new tasks. {\X} is able to achieve fairness and learn new tasks by controlling the rate-distortion function of {the} learned representations. Our empirical evaluations show that {\X} is able to make fair decisions while achieving high performance on the target task, outperforming {several} baselines. 
\end{abstract}

\section{Introduction}
An increasing number of organizations are leveraging machine learning solutions for making decisions in critical applications like hiring \citep{dastin2018amazon}, criminal recidivism \citep{larson2016we}, etc. Machine learning systems can often rely on a user's demographic information, like gender, race, and age (\textit{protected attributes}), encoded in their representations \citep{elazar2018adversarial} to make decisions, resulting in biased outcomes against certain demographic groups \citep{mehrabi2019survey, shah2019predictive}. Numerous works try to achieve {fairness through unawareness}  \citep{apfelbaum2010blind} by debiasing model representations from protected attributes \citep{blodgett2016demographic, elazar2018adversarial, elazar2021amnesic, chowdhury2022learning}. However, these techniques are only able to remove in-domain spurious correlations and fail to generalize to new data distributions \citep{barrett-etal-2019-adversarial}. For example, let us consider a fair resume screening system that was trained only on resumes of software engineering roles. The system may not remain fair while screening for roles like sales or marketing, where the gender {demographic distribution may be different}. Similarly, a fair system also needs to be robust to {shifts in data distribution}  (e.g. new applicants may report scores on specific tests that didn't appear in the training data) and {task changes} (e.g. resumes being screened for new roles like social media manager). {In such cases, it is not always practical to retrain  the system from scratch every time new data comes in  because of the resources and environmental impact associated with training modern machine learning systems.}

\begin{figure}[t!]
    \centering
    \includegraphics[width = 0.38\textwidth]{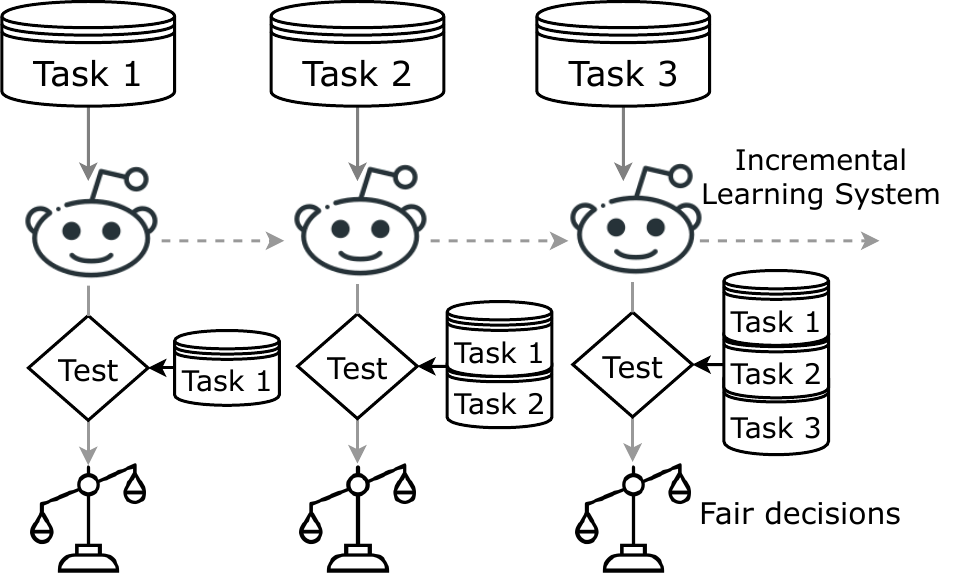}
    \caption{{Illustration of a fair representation learning system in an incremental setting. The system is expected to make fair decisions while incrementally learning new tasks.}}  
    \label{fig:intuition}
\end{figure}

{Previous works focused on improving the robustness of fair learning models by considering shifts in data distribution.  These involve learning fair models under covariate shift \cite{rezaei2021robust, singh2021fairness} or for streaming data \cite{zhang2021farf, zhang2019faht}, but these  systems do not incrementally  learn new tasks.}
In this work, we introduce the problem of \textit{learning fair representations in an incremental learning setting}. In this setting, data from new tasks, with different underlying demographic distributions, pour in at consecutive training stages and the system has to perform well on all tasks seen so far while making fair decisions (see Figure~\ref{fig:intuition}). This setup is quite similar to incremental learning \citep{rebuffi2017icarl}. However, most works in incremental learning literature focus on target task performance without considering the fairness of their predictions. 

To address this problem, we propose a representation learning system -- \textbf{F}airness-\textbf{a}ware \textbf{I}ncremental \textbf{R}epresentation \textbf{L}earning ({\X}). At its core, {\X} uses an adversarial debiasing setup for removing demographic information by controlling the 
  number of bits (\textit{rate-distortion})  required to encode  the learned representations  \citep{yu2020learning, ma2007segmentation}. We leverage this debiasing setup for incremental learning using an exemplar-based approach, by retaining a small set of representative samples from previous tasks, 
to prevent forgetting. Empirical evaluations show that {\X} outperforms baseline incremental learning systems 
in fairness metrics while successfully learning target task information. Our key contributions are:

\begin{itemize}
\item We propose {\X}, a representation learning system that learns fair representations, while incrementally learning new tasks, by controlling their rate-distortion function.
\item We show using empirical evaluations that {\X} outperforms baseline incremental learning systems in making fair decisions while performing well on the target task. 
\item We also perform extensive analysis experiments to investigate the functioning of  {\X}.
\end{itemize}

\section{Related Work}
In this section, {we discuss some of the prior works on fairness in varying setups and incremental learning.}

\noindent\textbf{Fair Representation Learning}. \citet{zemel2013learning} introduced the problem of learning fair representations as an optimization task. Following works~\citep{zhang2018mitigating, li2018towards, elazar2018adversarial, 
 basu-roy-chowdhury-etal-2021-adversarial} leveraged an adversarial framework~\citep{goodfellow2014generative} to achieve fairness, where a discriminator tries to extract demographic information from intermediate representations while performing prediction. Different from these, \cite{bahng2020learning} proposed to learn fair representations, without using protected attribute annotation, by making representations uncorrelated with ones retrieved from a biased classifier. However, these techniques require a target task 
at hand and are often difficult to train~\citep{elazar2018adversarial}. Another line of work introduced by \cite{bolukbasi2016man}, focuses on debiasing representations independent of a target task. These approaches~\citep{ravfogel2020null, bolukbasi2016man} iteratively identify subspaces that encode protected attribute information, and project vectors onto their corresponding nullspaces. Another line of work~\cite{fairfil, dixon2018measuring}, use counterfactual data augmentation approaches to debias sentence embeddings. Recently, \citet{chowdhury2022learning} proposed a debiasing framework that makes representations from  same protected attribute class  uncorrelated by maximizing their rate-distortion function. 
Despite showcasing promise 
in a single domain, these frameworks fail to remain fair for out-of-distribution data~\citep{barrett-etal-2019-adversarial}.

\noindent\textbf{Fairness under distribution shift}. {Several works \cite{rezaei2021robust, singh2021fairness} have investigated the robustness of fair classifiers under covariate shift. These works identify conditions where fairness can be sustained given shifts in data and label distribution. Efficacy of fair classifiers has also been studied in online settings \cite{zhang2021farf, zhang2019faht}, where the data distribution continually evolves depending on the input data stream. However, both lines of work consider a fixed task description at initiation and do not learn new tasks while training.} 

\noindent\textbf{Incremental Learning}. \citet{li2017learning} introduced the task of incremental learning and  proposed a dynamic architecture leveraging a knowledge distillation loss to prevent catastrophic forgetting \citep{mccloskey1989catastrophic}.
Since then, works on incremental learning can be classified into three broad categories: (a) \textit{Regularization-based} approaches~\citep{li2017learning, kirkpatrick2017overcoming, zenke2017continual, castro2018end, chan2021redunet} use a penalty measure to ensure model parameters crucial for previous tasks do not change abruptly; (b) \textit{Dynamic architecture-based} approaches~\citep{long2015learning, rusu2016progressive, li2019learn} introduce new task-specific parameters to prevent interference with parameters from previous tasks. These architectures grow linearly with the number of tasks having a heavy memory footprint; 
(c) \textit{Exemplar-based} approaches~\citep{rebuffi2017icarl, chaudhry2018efficient, chaudhry2019tiny, tong2022incremental} maintain a small memory of representative samples from previous tasks and replay them to prevent catastrophic forgetting. 
Our  framework {\X} is similar to that of \citet{tong2022incremental}, as we also control the rate-distortion of learned representations. However, we also consider the fairness of the predictions  by ensuring protected information does not get encoded in the representations.

\section{Background}
In this section, we discuss the fundamental concepts of rate-distortion theory that form the building blocks of our framework, {\X}. 

\noindent\textbf{Rate Distortion}. In information theory \citep{cover1999elements}, the compactness of a distribution is measured by their \textit{coding length} --  number of   bits required to encode it. In lossy data compression, a set of vectors  $Z = \{z_1, \ldots, z_n\} \in \mathbb{R}^{n \times d}${,} sampled from a distribution $P(Z)${,} is encoded using a coding scheme,  such that the transmitted vectors $\{\hat{z}_i\}^n_{i=1}$ can be recovered up to a distortion $\epsilon$. 
The minimal number of bits  required per vector to encode the sequence $Z$ is defined by the \textit{rate-distortion} function $R(Z, \epsilon)$. The optimal $R(Z, \epsilon)$ for vectors $Z$  sampled from a  multivariate Gaussian  $\mathcal{N}(0, \Sigma)$  is:
\begin{equation}
    R(Z, \epsilon) = \frac{1}{2} \log_2 \det \big(I + \frac{d}{n\epsilon^2}ZZ^T\big)
    \label{eqn:rate}
\end{equation}

\noindent where $n$ is the number of vectors and $d$ is the dimension of individual vectors. 
{Equation~\ref{eqn:rate} provides a tight bound  even in cases where the underlying distribution $P(Z)$ is degenerate \citep{ma2007segmentation}.}


{In general scenarios, e.g. image representations for multi-label classification, the vector set $Z$ can arise from a mixture of class distributions. 
In such cases, the overall rate-distortion function can be computed {by splitting} the vectors into  multiple subsets: $Z = Z^1 \cup Z^2 \ldots \cup Z^k$, where $Z^j$ is the subset from the $j$-th distribution. We can then compute $R(Z^j, \epsilon)$ (Equation~\ref{eqn:rate}) for each subset. 
To facilitate this computation, we leverage a global membership matrix $\Pi = \{\Pi_j\}^k_{j=1}$, which is a set of $k$ matrices encoding membership information in each subset. The membership matrix for a subset $Z^j$ is a diagonal matrix defined as:}
$\Pi_j = \mathrm{diag}(\pi_{1j}, \pi_{2j}, \ldots, \pi_{nj}) \in \mathbb{R}^{n \times n}$, 
{where $\pi_{ij} \in [0, 1]$ is the probability that $z_i$ belongs to $Z^j$. The matrices satisfy the following constraints: $\sum_j \Pi_j = I_{n \times n}$, $\sum_j\pi_{ij} = 1$, $\Pi_j \succeq 0$. The optimal number of bits to encode $Z$ is given as:}

\begin{equation*}
    R_c(Z, \epsilon | \Pi)  = \sum\limits_{j=1}^k \frac{\mathrm{tr}(\Pi_j) }{2n} \log_2 \det \big(I + \frac{d}{\mathrm{tr}(\Pi_j)\epsilon^2}Z\Pi_jZ^T\big)
\end{equation*}

{The expected number of vectors in a subset $Z^j$ is $\mathrm{tr}(\Pi_j)$ and the corresponding covariance  is $\mathrm{cov}(Z_j) = \frac{1}{\mathrm{tr}(\Pi_j)} Z\Pi_j Z^T$. For multi-class data, a vector $z_i$ can only be a member of a single class, {we restrict} $\pi_{ij} = \{0, 1\}$ and the covariance matrix for $j$-th subset is ${Z^j}(Z^j)^T$. } 

\noindent \textbf{Maximal Coding Rate (MCR\textsuperscript{2})}.\label{sec:mcr2} \citet{yu2020learning} introduced a {classification} framework by learning discriminative representations using the rate-distortion function. Given $n$ input samples $X = \{x_i\}^n_{i=1}$ belonging to $k$ distinct classes, their representations $Z = \{z_i\}^n_{i=1}$ are obtained using a deep network $f_{\theta}(x)$. The network parameters (${\theta}$) are learned by maximizing a representation-level objective using rate-distortion
called maximal coding rate (MCR\textsuperscript{2}): 

\begin{equation}
    \max_{\theta} \Delta R(Z, \Pi) = R(Z, \epsilon) - R_c(Z, \epsilon|\Pi)
    \label{eqn:mcr}
\end{equation}

where $\Pi$ captures the class label information. 
To have discriminative representations, same class representations should resemble each other while being different from representations from other classes. This can be achieved by maximizing the overall volume $R(Z, \epsilon)$ and compressing representations within each class by minimizing $R_c(Z, \epsilon| \Pi)$. We provide further details in Appendix~\ref{sec:mcr-ill}.

\section{Fairness-aware Incremental Representation Learning ({\X})}

\subsection{Debiasing Framework} 
\label{sec:debiasing-fram}

We present a novel adversarial debiasing framework that controls the rate-distortion function of the learned representations. {We use rate-distortion in this debiasing framework as it is amenable to incremental learning}.

Figure~\ref{fig:debiasing-framework} illustrates our proposed adversarial framework. It consists of a feature encoder $\phi$ and a discriminator  $D$. The feature encoder takes as input a data point $x$ and generates representations $z = \phi(x)$. 
Its goal is to learn representations that are discriminative for the target attribute $\mathbf y$ and not informative about protected attribute $\mathbf g$. 
The discriminator network takes as input the representations produced by feature encoder $z$ and generates $z' = D(z)$. Its goal is to extract protected attribute $\mathbf g$ information from $z'$.
The discriminator is trained by maximizing the MCR\textsuperscript{2} objective function: 
\begin{equation}
    \max_{D}  \Delta R(Z', \Pi^{\mathbf g}) = R(Z', \epsilon) - R_c(Z', \epsilon|\Pi^{\mathbf g})
\end{equation}

\noindent where $\Pi^{\mathbf g}$ is the membership matrix encoding the protected attribute information. The encoder is trained by optimizing the given objective function:

\begin{figure}[t!]
    \centering
    \includegraphics[width = 0.37\textwidth]{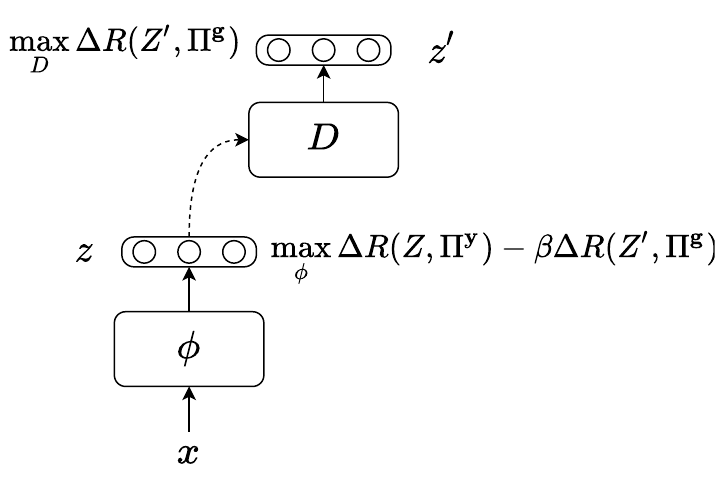}
    \caption{{Workflow of our debiasing framework. The discriminator tries to extract protected attribute information by optimizing $\Delta R(Z', \Pi^{\mathbf g})$. The feature encoder tries to learn discriminative representations for the target task ($\mathbf y$) using MCR\textsuperscript{2} objective while minimizing the discriminator loss.}}
    \label{fig:debiasing-framework}
\end{figure}

\begin{equation}
    \max_{\phi} \Delta R(Z, \Pi^{\mathbf y}) - \beta\Delta R(Z', \Pi^{\mathbf g})
    \label{eqn:enc}
\end{equation}

\noindent where $\Pi^{\mathbf y}$ is the membership matrix encoding the target attribute information and $\beta$ is a hyperparameter. 
Empirically we observed that the proposed debiasing framework is competitive with other debiasing setups in non-incremental learning settings (Section~\ref{sec:bmnist-results}). 

{\X}'s debiasing framework leverages the MCR\textsuperscript{2} objective  (Equation~\ref{eqn:mcr}) for classification. MCR\textsuperscript{2} objective, by itself, is not amenable to incremental learning for reasons discussed below. \citet{yu2020learning} showed that using MCR\textsuperscript{2} it is possible to learn representations with low-dimensional orthogonal subspaces corresponding to each class. However, naively maximimizing the MCR\textsuperscript{2} objective  results in representations spanning the complete feature space (an $\mathbb{R}^d$-dimensional feature space can accommodate a maximum of $d$ orthogonal subspaces). This is not ideal for incremental learning as representations from new classes cannot be accommodated in the same feature space. For incremental learning, representations learned at a given training stage should be compact and not span the entire feature space. In {\X}, we empirically  observe that the feature spaces learned at each training stage are compact (Section~\ref{sec:compact}). 
This happens because while learning discriminative representations using the MCR\textsuperscript{2} objective ($\Delta R(Z, \Pi^{\mathbf y})$), the encoder also tries to remove protected information by minimizing  $\Delta R(Z', \Pi^{\mathbf g})$ (Equation~\ref{eqn:enc}). Minimizing $\Delta R(Z', \Pi^{\mathbf g})$  {makes representations from different protected classes similar, resulting in a compact feature space}. The $\Delta R(Z', \Pi^{\mathbf g})$ term acts as a natural regularizer to the MCR\textsuperscript{2} objective, and prevents the learned representations from expanding in an unconstrained manner, making them suitable for incremental learning. Next, we  discuss how we extend this debiasing framework to incremental learning in the following section.


\subsection{Incremental Learning}
\label{sec:IL}

For incremental learning, we use an \textit{exemplar-based} approach \citep{rebuffi2017icarl, chaudhry2018efficient, chaudhry2019tiny}. We store a small set of exemplars from old tasks $\mathcal X_{old} = \{\mathcal {X}_{old}^1, \ldots, \mathcal X_{old}^{m}\}$, where $m$ is the number of target classes $(m = c(t-1) < k)$ the system has encountered so far (each training step introduces $c$ target classes, $k$ is the total number of classes). At training stage $t$, we have a set of new data samples $\mathcal X_{new}$ and exemplar set $\mathcal X_{old}$  ($\mathcal X_{old} = \emptyset$ at $t=0$).
The goal of our system is to learn discriminative representations w.r.t $\mathbf{y}$ for $\mathcal X_{new}$ while retaining the old representation subspaces of $\mathcal X_{old}$. To ensure fairness, the system also needs to learn representations that are oblivious to the protected attribute $\mathbf g$ for both $\mathcal X_{new}$ and $\mathcal X_{old}$. We will refer to the representations 
for the old and new data as $Z_{old} = \phi(\mathcal{X}_{old})$ and $Z_{new} = \phi(\mathcal{X}_{new})$ respectively.

\noindent \textbf{Discriminator}. In the incremental learning setup, the discriminator  tries to extract protected attribute information for $\mathcal X_{new}$. This is achieved by maximizing $\Delta R(Z'_{new}, \Pi^{\mathbf g}_{new})$, where $Z'_{new} = D(\phi(\mathcal{X}_{new}))$, and $\Pi^{\mathbf g}_{new}$ encodes protected attribute $\mathbf{g}$ information for $\mathcal{X}_{new}$.

\noindent \textbf{Feature encoder}. 
The objective of the feature encoder is to learn fair representations that are discriminative for both old and new tasks. To achieve this, the  system should have the following properties:

\begin{enumerate}[label=(\alph*)]
    \item The system should learn  representations  for $\mathcal{X}_{new}$ that are informative about $\mathbf y$. This can be achieved by learning discriminative representations for $\mathcal{X}_{new}$ by \textit{maximizing} the MCR\textsuperscript{2} objective: $\Delta R(Z_{new}, \Pi^{\mathbf y}_{new})$.
    \item The system should not reveal protected information and learn fair representations for $\mathcal{X}_{new}$. This is achieved by \textit{minimizing} the discriminator loss $\Delta R(Z'_{new}, \Pi^{\mathbf g}_{new})$ (Equation~\ref{eqn:enc}).
    \item The system should retain knowledge about old tasks encountered in previous training stages. {\X} maintains an exemplar set $\mathcal X_{old}$ and tries to retain the subspace structure learned for these samples. To ensure that encoder $\phi_t$ at training stage $t$ retains the subspace structure of old representations, we \textit{minimize} the function:
    
    \begin{equation}
    \begin{aligned}
        \Delta R(Z_{old}, \bar{Z}_{old}) 
        = \sum_{i=1}^m &\Delta R(Z^i_{old}, \bar{Z}^i_{old}) \\
        = \sum_{i=1}^m R(Z^i_{old} \cup \bar{Z}^i_{old}) &- \frac{1}{2} \left[R(Z^i_{old}) + R(\bar{Z}^i_{old})\right]
    \end{aligned}
    \end{equation}
    
    where $\bar{Z}_{old} = \phi_{t-1}(\mathcal X_{old})$ are exemplar representations  obtained using the encoder at the previous training stage ($t-1$), {and} $Z_{old}^j$ are exemplar representations from the $j$-{th} target class. $\Delta R(Z^j_{old}, \bar{Z}^j_{old})$ measures the similarity between the representation sets $Z^j_{old}$ and $\bar{Z}^j_{old}$ by computing the difference in the number of bits required to encode them jointly and separately. 
    
    \item The system should learn fair representations for $\mathcal{X}_{old}$. This is achieved by \textit{minimizing} the discriminator loss for the exemplars $\Delta R(Z'_{old}, \Pi^{\mathbf g}_{old})$ (Equation~\ref{eqn:enc}).
\end{enumerate}

The overall objective function that the encoder optimizes in the incremental learning setup:

\begin{equation}
    \begin{aligned}
    \max_{\phi} \underbrace{\Delta R(Z_{new}, \Pi^{\mathbf y}_{new})}_{\mathrm{(a)}} - \beta\underbrace{\Delta R(Z'_{new}, \Pi^{\mathbf g}_{new})}_{\mathrm{(b)}} \\- \gamma\underbrace{\Delta R(Z_{old}, \bar{Z}_{old})}_{\mathrm{(c)}} - \eta\underbrace{\Delta R(Z'_{old}, \Pi^{\mathbf g}_{old})}_{\mathrm{(d)}} 
    \end{aligned}
    \label{eqn:enc-inc}
\end{equation}

\noindent where $Z_{old}' = D(Z_{old})$, $\Pi^{\mathbf y}_{new}$ is the membership matrix encoding target class labels for $\mathcal X_{new}$, $\Pi^{\mathbf g}_{new}$ and $\Pi^{\mathbf g}_{old}$ encode protected class labels for $\mathcal X_{new}$ and $\mathcal X_{old}$ respectively. 
In the following section, we discuss the selection of representative samples $\mathcal{X}_{old}$ from old classes.

\subsection{Exemplar Sample Selection}
\label{sec:exemplar}

As discussed in previous section, we maintain exemplars $\mathcal{X}_{old} = \{\mathcal{X}_{old}^1, \ldots, \mathcal{X}_{old}^m\}$ belonging to $m$ classes, which is useful for retaining information from previous tasks. For each class, we select $r$ (where $r\ll \lvert \mathcal{X}^i\rvert)$ samples $\mathcal{X}_{old}^i \sim \mathcal{X}^i$ by using one of the following sampling techniques:

   \noindent\textbf{Random Sampling}. We randomly select $r$ samples from each class set $\mathcal{X}^i_{old} {\stackrel{\mathclap{r}}{\sim}} \mathcal{X}^i$. 
   
    \noindent\textbf{Prototype Sampling}. We use prototype sampling \citep{tong2022incremental} for selecting representative samples for each class. The detailed pseudo-code is presented in Algorithm~\ref{alg:prototype}. In this technique, we compute the top $k$ eigenvectors for the set of representations for each class $Z^i_t = \phi(\mathcal{X}^i_t)$ at training stage $t$. For each eigenvector, we select $r/k$ data samples ($\mathcal{X}_{old}^t$) with the highest similarity scores (line~\ref{alg:line:sim}). The selected samples are added to $\mathcal{X}_{old}$. 
    
      \begin{algorithm}[t!]
        \caption{Prototype Sampling}
        \begin{algorithmic}[1]
            \State \textbf{Input}: $Z_t = \{\phi(\mathcal{X}_t^1), \ldots, \phi(\mathcal{X}_t^c))\}$ {representations of $c$ classes at training stage $t$, reservoir of old  samples $\mathcal{X}_{old}$.}
            \State $\mathcal{X}_{old}^t = \emptyset$ \Comment{exemplars for training stage $t$}
            \For{$i = 1, \ldots, c $}
                \State $V^i = \texttt{PCA}(Z_t^i)$ \Comment{where $Z_t^i = \mathcal{X}_t^i$}
                \State $V^i_k = [v_1, \ldots, v_k] \sim \text{top-}k(V^i)$ \Comment{top-$k$ eigen vectors selected based on singular values} 
                \For{$j = 1, \ldots, r$}
                    \State $s = v_i^T Z_{t}^i$ \Comment{similarity scores}\label{alg:line:sim}
                    \State $\mathcal{X}_{old}^i \sim \text{top-}q(\mathcal{X}^i_t)$ \Comment{select top $q = \frac{r}{k}$ samples based on similarity scores $s$}
                    \State $\mathcal{X}_{old}^t = \mathcal{X}_{old}^t \cup \mathcal{X}_{old}^i$ 
                \EndFor
            \EndFor
        \State $\mathcal{X}_{old} = \mathcal{X}_{old} \cup \mathcal{X}_{old}^t$ \Comment{add to exemplar set}
        \State \Return $\mathcal{X}_{old}$
        \end{algorithmic}
        \label{alg:prototype}
      \end{algorithm}

    \noindent\textbf{Submodular Optimization}. We use submodular optimization \citep{krause2014submodular} to select representative samples that summarize features of a set. Submodular optimization focuses on set functions which have the diminishing return property. Formally, a submodular function $f$ satifies the property: $f(Z \cup \{s\}) - f(Z) \geq f(Y \cup \{s\}) - f(Y)$ , where $Z \subseteq Y \subseteq S$,  $s \in S$, and $s \not\in Y$. 
    
    We construct a submodular function computed using representations $Z$ that capture their diversity. 
     We select $r$ samples that maximizes $f$. Specifically, we use the facility location algorithm \citep{frieze1974cost}, which selects $r$ representative samples from a set $Z$ with $n$ elements ($n > r$). For any subset $S \subseteq Z$, the submodular function $f$ is: $f(S) = \sum_{z \in Z} \max_{s \in S} \text{sim}(s, z)$,
    \noindent where $\text{sim}(\cdot, \cdot)$ is the similarity measure between $s$ and $z$. In our experiments, $Z$ is the set of data representations 
    and we use euclidean distance as our similarity measure $\text{sim}(s, z) = -\lvert\lvert s - z \rvert\rvert^2_2$. 

\section{Evaluation}

In this section, we discuss the datasets, experimental setup, and metrics used for evaluating {\X}.  
 {Additional} details of our experimental setup can be found in Appendix~\ref{sec:impl}. Our implementation of {\X} is publicly available at \texttt{https://github.com/brcsomnath/FaIRL}. 

\subsection{Datasets}
\label{sec:dataset}
We tackle the problem of fairness in an incremental learning setup, where there are no existing benchmarks.\footnote{Most fairness datasets have target attributes with only 2 classes (along with a binary protected attribute), making them unsuitable for evaluating incremental learning.} We perform evaluations by re-purposing existing datasets.

\noindent\textbf{Biased MNIST}. We follow the setup of \citep{bahng2020learning} to generate a synthetic dataset using MNIST \citep{lecun1998gradient}, by making the background colors highly correlated with the digits. In the training set, the digit category (\textit{target attribute}) is associated with a distinct background color (\textit{protected attribute}) with probability $p$ or a randomly chosen color with probability $1- p$. In the test set, each digit with assigned one of the 10 colors randomly. We  evaluate the generalization ability of {\X} for {$p = \{0.8, 0.85, 0.9, 0.95\}$}. We simulate incremental learning by providing the system access to 2 classes at each training stage (a total of 5 stages).

\begin{figure}[h!]
    \centering
    \includegraphics[width = 0.25\textwidth]{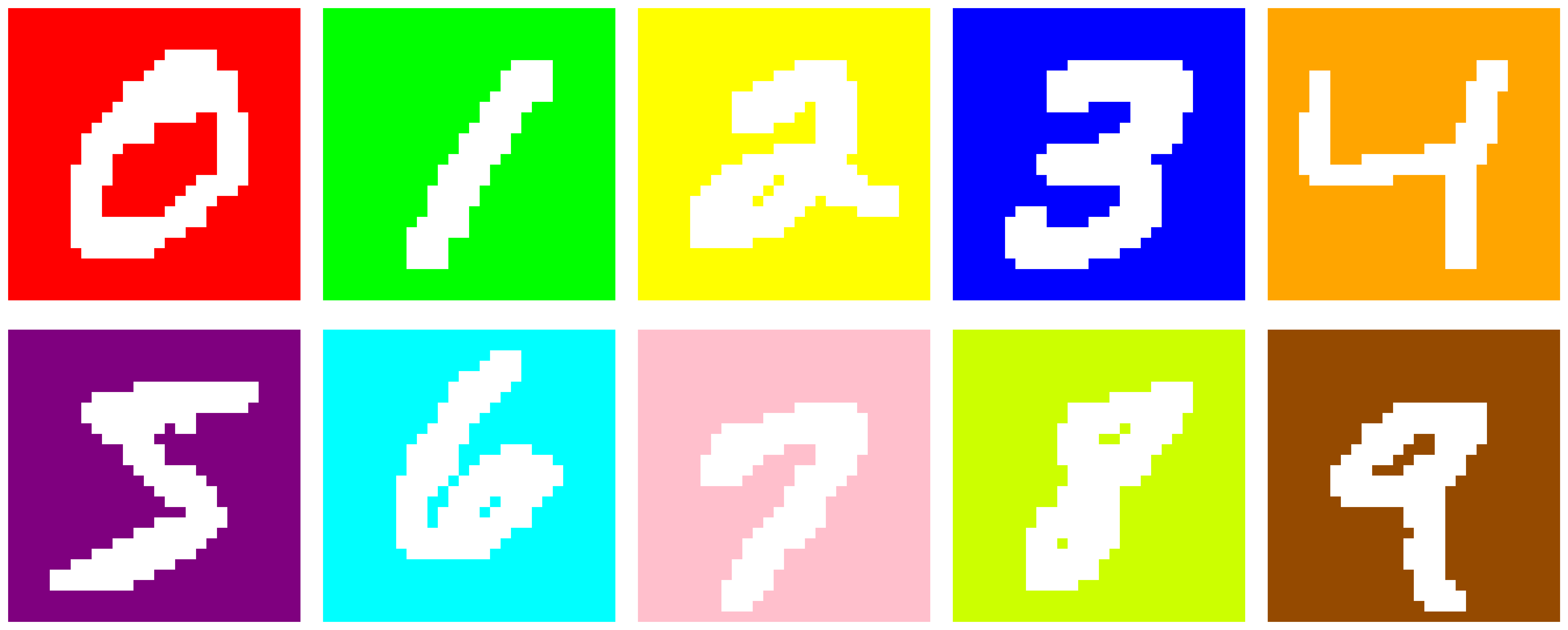}
    \caption{ {Representative samples from Biased MNIST dataset. We show an example from each class.}}
    \label{fig:biased-mnist}
\end{figure}

\noindent\textbf{Biography classification}. We re-purpose the  BIOS dataset~\citep{de2019bias} for incremental learning. BIOS contains biographies of people that are associated with a profession (\textit{target attribute}) and gender label (\textit{protected attribute}). There are 28 different profession categories and 2 gender classes.
The demographic distribution can vary vastly depending on the profession (e.g. `\textit{software engineer}' role is skewed towards men while the `\textit{yoga teacher}' role is most associated with females). The detailed demographic distribution is reported in Appendix~\ref{sec:data-stats}. 
In our setup, the system is presented with samples from 5 classes at each training stage (a total of 6 training stages).

\subsection{Baselines}
We compare {\X} with the following systems: 
    
\noindent$\bullet$ \textbf{Incremental Learning systems}. We report the performance of the {following} incremental systems: (a) {LwF} \citep{li2017learning} {is} a dynamic architecture with shared and task specific parameters, with additional parameters being incorporated incrementally for new tasks. LwF uses a knowledge distillation loss along with the current task loss to prevent catastrophic forgetting; (b) {Adversarial LwF} --  we introduced an adversarial head in LwF for fair incremental learning that tries to remove protected attribute information via gradient reversal; (c) {iCaRL} \citep{rebuffi2017icarl} is an exemplar-based approach that uses a knowledge distillation loss to learn representations. iCaRL uses a nearest class mean classifier for performing prediction.

\noindent$\bullet$ \textbf{Joint learning systems}. We report the performance of the following joint learning systems, where the system has access to the entire dataset in a single training stage:  (a) AdS \citep{basu-roy-chowdhury-etal-2021-adversarial} is an adversarial debiasing framework that maximizes the entropy of discriminator output. 
(b)  FaRM \citep{chowdhury2022learning} is a state-of-the-art  system for both constrained and unconstrained debiasing,  which performs debiasing by controlling the rate-distortion function of representations; (c) {\X} ({joint}).
    We report the performance of our framework when trained on full data.

\subsection{Metrics}
In this section, we discuss the metrics reported. For each metric, we report the average and the value achieved at the final training stage.

\noindent$\bullet$ \textbf{Target Accuracy}. 
We follow \citep{elazar2018adversarial, ravfogel2020null, basu-roy-chowdhury-etal-2021-adversarial} in evaluating the quality of the learned representations for target task ($\mathbf y$) by using a separate probing network. 

For Biased MNIST, a fair system would be able to generalize to the test set, therefore target accuracy helps measure the fairness of the system. A \textit{high accuracy} is desired in all settings.  For both datasets, we also report group fairness metrics discussed below.

\noindent$\bullet$ \textbf{Group Fairness Metrics}.\label{sec:metrics}  We evaluate the fairness of representations using the following metrics. A \textit{low score} on these metrics indicates a fairer system. 

    \noindent(a) \textbf{TPR-GAP}.  TPR-GAP \citep{de2019bias} computes the difference between true positive rates between two protected groups
    $\mathrm{Gap}_{\mathbf{g},y} = \mathrm{TPR}_{g,y} - \mathrm{TPR}_{\bar{g},y}$, 
    \noindent where $g, \bar{g}$ are possible values of the protected attribute. 
     \citep{biasbios2} proposed a single fairness score by computing the root mean square of $\mathrm{Gap}_{\mathbf{g}, y}$: $\mathrm{Gap}_\mathbf{g}^{\mathrm{RMS}} = \sqrt{{1/}{\lvert \mathcal{Y}\rvert} \sum_{y \in \mathcal{Y}} (\mathrm{Gap}_{\mathbf{g},y})^2}$, where $\mathcal Y$ is the target label set. 
    
    \noindent(b) \textbf{Demographic Parity (DP).} DP measures the difference in target prediction rate w.r.t to protected attribute $\mathbf g$. {Mathematically, it is expressed as:} 
    \begin{equation}
		\mathrm{DP} = \sum\limits_{y \in \mathcal{Y}}\lvert p(\hat{\mathbf y} = y|\mathbf{g} = g) -  p(\hat{\mathbf y} = y|\mathbf{g} = \bar{g})\rvert
    \end{equation}

\label{sec:fair-metrics}
\citet{zhao2019inherent} illustrated that there is an inherent tradeoff between the utility and fairness in fair representation learning, when $\mathbf y$ and $\mathbf g$ are correlated. Accordingly, in our experiments, we observe good fairness scores often result in poor target task performance and vice-versa. 

\begin{table*}[t!]
	\centering
	\resizebox{0.72\textwidth}{!}{
\begin{tabular}{@{}cl@{}| c c | c c|c c| c c}
    \toprule[1pt]
    	\multicolumn{2}{l@{~}|}{\multirow{2}{*}{\textbf{Method}}} & \multicolumn{2}{c|}{$p=0.8$} & \multicolumn{2}{c|}{$p=0.85$} & \multicolumn{2}{c|}{$p=0.9$} & \multicolumn{2}{c}{$p=0.95$} \Tstrut\\
     & & Last & Avg. & Last & Avg. & Last & Avg. & Last & Avg. \Bstrut\\
    \midrule[0.5pt]
    \parbox[t]{0.8mm}{\multirow{7}{*}{\rotatebox[origin=c]{90}{\small{\texttt \textbf{Incremental}}}}} &  LwF \citep{li2017learning} & 10.28 & 32.39 & 10.28 & 31.50 & 10.60 & 31.34 & 10.28 & 28.56\\
    & Adversarial LwF & 10.28 & 32.38 & 10.28 & 31.94 & 10.28 & 27.13 & 10.28 & 25.79\\
   &   iCaRL \citep{rebuffi2017icarl} & 62.81 & 79.20 & 58.44 & 72.41 & 51.08 & 70.82 & 47.48 & 69.86 \Bstrut\\
     \cmidrule[0.5pt]{2-10}
    &  {\X} (\texttt{random}) & \textbf{81.67} & \textbf{90.44} & \textbf{77.76} & \textbf{88.21} & 71.05 & 83.89 & \textbf{59.26} & \textbf{75.66}\Tstrut\\
    &  {\X} (\texttt{prototype}) & 80.73 & 89.82 & 77.18 & 87.69 & 70.95 & 83.47 & 57.79 & 75.28\\
    &  {\X} (\texttt{submod})  &  {80.52} & {89.84} & {77.62} & {88.01} & \textbf{72.24} & \textbf{84.38} & {57.94} & {73.48} \Bstrut\\
    \midrule[0.5pt]
    \parbox[t]{0.8mm}{\multirow{3}{*}{\rotatebox[origin=c]{90}{\small\texttt Joint}}} & {\X} (\texttt{joint}) & \light{88.08} & - & \light{85.64} & - & \light{81.94} & - & \light{68.85} & - \Tstrut\\
    & AdS \citep{basu-roy-chowdhury-etal-2021-adversarial} & \light{79.98} & - & \light{75.39} & - & \light{66.46} & - & \light{52.49} & -\\
    & {FaRM \citep{chowdhury2022learning} } & \light{92.44} & - & \light{90.54} & - & \light{82.55} & - & \light{57.09}\Bstrut\\
    \bottomrule[1pt]
\end{tabular}
}
	\caption{{Evaluation accuracy of incremental and joint learning systems on Biased MNIST dataset. Performance of joint learning systems are reported in \light{gray}. {\X} achieves the best performance among incremental learning baselines (shown in \textbf{bold}). In strongly correlated settings ($p=0.95$), {\X} is competitive with joint learning setups.}}
	\label{tab:bmnist}
\end{table*}
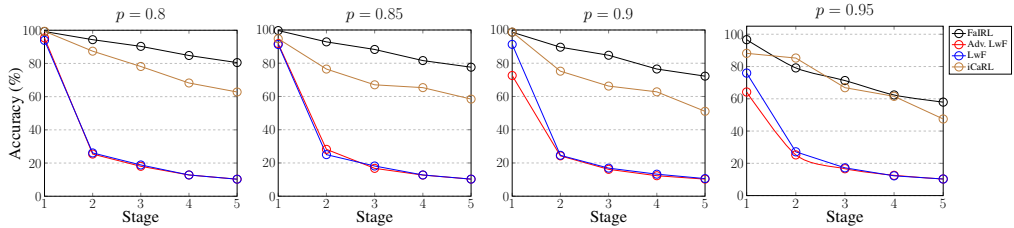
\begin{figure*}[t!]
	\centering
	\resizebox{0.18\textwidth}{!}{
		\begin{tikzpicture}
\pgfplotsset{
    scale only axis,
    tick label style={font=\huge},
}

\begin{axis}[
title={\Huge $p=0.8$},
  ymin=0, ymax=100,
  xmin=1, xmax=5,
  xtick={1, 2, 3, 4, 5},
  ytick={},
    ymajorgrids=true,
    grid style=dashed,
  xlabel={\Huge Stage},
  ylabel={\Huge Accuracy (\%)},
  legend pos=south west,
]

\addplot[mark=o,mark options={scale=2.5},black]
  coordinates{
(1, 99.2434988179669)
(2, 94.32282896319461)
(3, 90.30011606698723)
(4, 84.79481102656854)
(5, 80.52000000000001)
}; 

\addplot[mark=o,mark options={scale=2.5},red]
  coordinates{
    (1, 95.36643026004729)
    (2, 25.402934808756317)
    (3, 18.04012601558614)
    (4, 12.822751652737932)
    (5, 10.28)
}; 

\addplot[mark=o,mark options={scale=2.5},blue]
  coordinates{
    (1, 93.90070921985816)
    (2, 26.100553283617994)
    (3, 18.852594926214558)
    (4, 12.822751652737932)
    (5, 10.28)
}; 

\addplot[mark=o,mark options={scale=2.5},brown]
  coordinates{
    (1, 99.38534545898438)
    (2, 87.3947525024414)
    (3, 78.17940521240234)
    (4, 68.23001098632812)
    (5, 62.810001373291016)
}; 

\end{axis}
\end{tikzpicture}{\label{subfig:bmnist-0.8}}}
	\resizebox{0.17\textwidth}{!}{
		\begin{tikzpicture}
\pgfplotsset{
    scale only axis,
    tick label style={font=\huge},
}

\begin{axis}[
title={\Huge $p=0.85$},
  ymin=0, ymax=100,
  xmin=1, xmax=5,
  xtick={1, 2, 3, 4, 5},
  ytick={},
    ymajorgrids=true,
    grid style=dashed,
  xlabel={\Huge Stage},
  legend pos=south west,
]

\addplot[mark=o,mark options={scale=2.5},black]
  coordinates{
(1, 99.66903073286052)
(2, 92.8554245850373)
(3, 88.29381528768032)
(4, 81.60159660720969)
(5, 77.61999999999999)
}; 

\addplot[mark=o,mark options={scale=2.5},red]
  coordinates{
    (1, 91.77304964539007)
    (2, 28.19340870820303)
    (3, 16.763389156027192)
    (4, 12.710490208307346)
    (5, 10.28)
}; 

\addplot[mark=o,mark options={scale=2.5},blue]
  coordinates{
    (1, 91.25295508274232)
    (2, 25.01804185710849)
    (3, 18.17277400099486)
    (4, 12.78533117126107)
    (5, 10.28)
}; 

\addplot[mark=o, mark options={scale=2.5}, brown]
  coordinates{
    (1, 94.75177001953125)
    (2, 76.49747467041016)
    (3, 67.02039337158203)
    (4, 65.32368469238281)
    (5, 58.439998626708984)
}; 

\end{axis}
\end{tikzpicture}{\label{subfig:bmnist-0.85}}}
	\resizebox{0.17\textwidth}{!}{
		\begin{tikzpicture}
\pgfplotsset{
    scale only axis,
    tick label style={font=\huge},
}

\begin{axis}[
title={\Huge $p=0.9$},
  ymin=0, ymax=100,
  xmin=1, xmax=5,
  xtick={1, 2, 3, 4, 5},
  ytick={},
    ymajorgrids=true,
    grid style=dashed,
  xlabel={\Huge Stage},
  legend pos=south west,
]

\addplot[thin, mark=o,mark options={scale=2.5},black]
  coordinates{
(1, 98.67612293144208)
(2, 89.65600192446476)
(3, 84.81180567070137)
(4, 76.5248846201821)
(5, 72.24000000000001)
}; 

\addplot[thin, mark=o,mark options={scale=2.5},red]
  coordinates{
    (1, 72.62411347517731)
    (2, 24.272311763290835)
    (3, 16.166473221687946)
    (4, 12.32381189971311)
    (5, 10.28)
}; 

\addplot[ultra thin, mark=o,mark options={scale=2.5},blue]
  coordinates{
    (1, 91.34751773049645)
    (2, 24.60909309598268)
    (3, 16.896037141435916)
    (4, 13.25932393663465)
    (5, 10.6)
}; 

\addplot[thick, mark=o,mark options={scale=2.5}, brown]
  coordinates{
    (1, 98.77068328857422)
    (2, 75.22251892089844)
    (3, 66.2576675415039)
    (4, 62.791568756103516)
    (5, 51.08000183105469)
}; 

\end{axis}
\end{tikzpicture}{\label{subfig:bmnist-0.9}}}
	\resizebox{0.22\textwidth}{!}{
		\begin{tikzpicture}
\pgfplotsset{
    scale only axis,
    tick label style={font=\huge},
}

\begin{axis}[
title={\Huge $p=0.95$},
  ymin=0, ymax=105,
  xmin=1, xmax=5,
  xtick={1, 2, 3, 4, 5},
  ytick={},
    ymajorgrids=true,
    grid style=dashed,
  xlabel={\Huge Stage},
  legend pos=outer north east,
  legend cell align={left},
]

\addplot[smooth, mark=o,mark options={scale=2.5},black]
  coordinates{
(1, 96.64302600472813)
(2, 79.1436131825836)
(3, 71.29829215718786)
(4, 62.35499563427717)
(5, 57.940000000000005)
}; \addlegendentry{\Large {\X}}

\addplot[smooth, mark=o, mark options={scale=2.5}, red]
  coordinates{
    (1, 64.2080378250591)
    (2, 25.23454414241039)
    (3, 16.647322168794562)
    (4, 12.5608082823999)
    (5, 10.28)
}; \addlegendentry{\Large Adv. LwF}

\addplot[mark=o,mark options={scale=2.5},blue]
  coordinates{
    (1, 75.9338061465721)
    (2, 27.134953091171515)
    (3, 17.211076106781628)
    (4, 12.248970936759386)
    (5, 10.28)
}; \addlegendentry{\Large LwF}

\addplot[mark=o, mark options={scale=2.5}, brown]
  coordinates{
    (1, 88.17967224121094)
    (2, 85.25379180908203)
    (3, 66.88774871826172)
    (4, 61.51927185058594)
    (5, 47.47999954223633)
}; \addlegendentry{\Large iCaRL}

\end{axis}
\end{tikzpicture}{\label{subfig:bmnist-0.95}}}
	\caption{{Test accuracy at different training stages of {\X} and baseline incremental learning systems on Biased MNIST dataset. We observe that {\X} significantly outperforms baseline approaches in all setups.}}
	\label{fig:loss-evolution-bmnist}
\end{figure*}

\subsection{Results: Biased MNIST}
\label{sec:bmnist-results}

In Table~\ref{tab:bmnist}, we report the performance of {\X} and baseline approaches on Biased MNIST dataset. 
For this dataset, high target accuracy also implies fair decisions as the training sets are biased. 
We observe that {\X}  outperforms the  incremental learning baselines in all settings (different values of $p$). 
{\X} with prototype and submodular exemplar selection approaches  slightly fall behind random sampling. We believe that as the class samples are skewed towards a color, these sampling approaches may have ended up selecting instances based on their color instead of the digit information. 
It is also interesting to note that {\X} ({joint}) is competitive with other state-of-the-art approaches AdS and FaRM, outperforming them when the color and digit information are strongely correlated ($p=0.95$). In this settings ($p=0.95$), {\X} even in the incremental learning setting outperforms joint learning baselines. This shows that {\X} is able to learn robust  representations in challenging scenarios where the bias is highly correlated with the target task. We report the fairness metrics in Appendix~\ref{sec:addl-bmnist} for completeness. 

In Figure~\ref{fig:loss-evolution-bmnist}, we report the performance of incremental learning systems at various training stages. We observe that LwF suffers from catastrophic forgetting, achieving near random performance in the final stages. Adversarial LwF achieves a similar performance to LwF. We believe that the adversarial head doesn't provide an added advantage over LwF because it may encounter unseen classes of protected attribute (colors) at  later training stages. iCaRL and {\X} do not suffer from catastrophic forgetting, and in all settings {\X} consistently outperforms other baselines.

\subsection{Results: Biography Classification}
We present the results of {\X} on Biography classification in Table~\ref{tab:results-bios}. 
We observe that LwF-based systems achieve poor target performance due to catastrophic forgetting. However, as most of their predictions are incorrect these systems end up with good scores on fairness metrics. 
Adversarial LwF performs slightly better than LwF in terms of target accuracy.  iCaRL achieves the best target accuracy but performs the worst on fairness metrics. {\X} provides a good balance between the two traits -- achieving target accuracy close to iCaRL while significantly improving the fairness metrics.

\begin{table}[t!]
	\centering
	\resizebox{0.48\textwidth}{!}{
	\begin{tabular}{@{}cl@{}| c c   c c  c c} 
			\toprule[1pt]
			 \multicolumn{2}{l@{~}|}{\multirow{3}{*}{\textbf{Method}}} & \multicolumn{2}{c}{\multirow{2}{*}{Accuracy ($\uparrow$)}} & \multicolumn{4}{c}{Fairness} \Tstrut\\
			 & & \multicolumn{2}{c}{} & \multicolumn{2}{c}{DP ($\downarrow$)} & \multicolumn{2}{c}{$\mathrm{Gap}^{\mathrm{RMS}}_{\mathbf g}$ ($\downarrow$)}\\
			 & & Last & Avg. & Last & Avg. & Last & Avg.  \Bstrut\\
			\midrule[0.5pt]
			\parbox[t]{0.8mm}{\multirow{7}{*}{\rotatebox[origin=c]{90}{\small{\texttt Incremental}}}} & LwF \citep{li2017learning} & 17.90 & 52.09 & 0.25 & 0.30 & 0.045 & 0.015\\
			& Adversarial LwF & 21.07 & 54.21 & 0.31 & 0.36 & 0.185 & 0.047\\
			& iCaRL \citep{rebuffi2017icarl} & 97.72 & 99.08  & 0.45 & 0.37 & 0.095 & 0.052\\
			\cmidrule[0.5pt]{2-8}
			& {\X} (\texttt{random}) & 95.11 & 97.49 & 0.42 & 0.35 & 0.061 & 0.032\Tstrut\\
			& {\X} (\texttt{prototype}) & 93.93 & 96.82 & 0.40 & 0.34 & 0.050 & 0.031\\
			& {\X} (\texttt{submodular}) & 94.37 & 97.43 & 0.41 & 0.35 & {0.044} & 0.022 \Bstrut\\
			\midrule[0.5pt]
			\parbox[t]{0.8mm}{\multirow{3}{*}{\rotatebox[origin=c]{90}{\small\texttt Joint}}} & {\X} (\texttt{joint}) & \light{98.45} & - &  \light{0.43} & - & \light{0.048} & - \Tstrut\\
			& {AdS \citep{basu-roy-chowdhury-etal-2021-adversarial}} & \light{99.90} & - & \light{0.45} & - & \light{0.003} & - \\
			& {FaRM (Chowdhury et al. 2022) \hspace{0.05cm}} & \light{99.90} & -  & \light{0.42} & - & \light{0.002} & -\Bstrut\\
			\bottomrule[1pt]
	\end{tabular}
}
	\caption{{Target accuracy and fairness metrics achieved by {\X} and other baseline approaches on Biographies dataset. {\X} achieves a good balance between target accuracy and fairness metrics.}}
	\label{tab:results-bios}
\end{table}

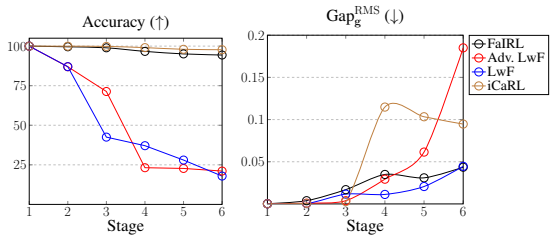
\begin{figure}[t!]
	\centering
	
	\resizebox{0.17\textwidth}{!}{
		\begin{tikzpicture}
\pgfplotsset{
    scale only axis,
    tick label style={font=\huge},
}

\begin{axis}[
title={\Huge Accuracy ($\uparrow$)},
  ymin=00, ymax=105,
  xmin=1, xmax=6,
  xtick={1, 2, 3, 4, 5, 6},
  ytick={25, 50, 75, 100},
    ymajorgrids=true,
    grid style=dashed,
  xlabel={\Huge Stage},
  legend pos=south west,
]

\addplot[smooth, mark=o,mark options={scale=2.5},black]
  coordinates{
(1, 99.981439298068)
(2, 99.56776860542551)
(3, 98.93160718085404)
(4, 96.6671201197116)
(5, 95.07260564529287)
(6, 94.36835629424934)
}; 

\addplot[mark=o,mark options={scale=2.5},red]
  coordinates{
(1, 99.98987598076436)
(2, 86.96395139909784)
(3, 71.32473990122405)
(4, 23.184705377604303)
(5, 22.725974873551966)
(6, 21.068237287451844)
}; 

\addplot[mark=o,mark options={scale=2.5},blue]
  coordinates{
(1, 100)
(2, 87.00918491713471)
(3, 42.513970859977823)
(4, 37.09071502568986)
(5, 28.01435797030511)
(6, 17.9004377029671)
}; 

\addplot[smooth, mark=o,mark options={scale=2.5},brown]
  coordinates{
(1, 100.0)
(2, 99.94722747802734)
(3, 99.80625915527344)
(4, 99.04460906982422)
(5, 97.98294830322266)
(6, 97.72071838378906)
}; 

\end{axis}
\end{tikzpicture}{\label{subfig:bios-acc}}}
	\resizebox{0.235\textwidth}{!}{
		\begin{tikzpicture}
\pgfplotsset{
    yticklabel style={
        /pgf/number format/fixed,
        /pgf/number format/precision=2
    },
    scaled y ticks=false,
    scale only axis,
    tick label style={font=\huge},
}

\begin{axis}[
title={\Huge $\text{Gap}_{\mathbf g}^{\mathrm{RMS}}$ ($\downarrow$)},
  ymin=00, ymax=.2,
  xmin=1, xmax=6,
  xtick={1, 2, 3, 4, 5, 6},
  ytick={0.05, 0.1, 0.15, 0.2},
    ymajorgrids=true,
    grid style=dashed,
  xlabel={\Huge Stage},
  legend pos=outer north east,
  legend cell align={left},
]

\addplot[smooth, mark=o,mark options={scale=2.5},black]
  coordinates{
(1, 0.00027992024875125583)
(2, 0.0038108560735470877)
(3, 0.016908815238406726)
(4, 0.03492383769200393)
(5, 0.030915474346487653)
(6, 0.04353209512557307)
}; \addlegendentry{\huge {\X}}

\addplot[smooth, mark=o,mark options={scale=2.5},red]
  coordinates{
(1, 0.00022102955926734655)
(2, 0.0008526825479292658)
(3, 0.0038094280380789485)
(4, 0.029308193765319818)
(5, 0.06145817196874671)
(6, 0.18500822639084297)
}; \addlegendentry{\huge Adv. LwF}

\addplot[smooth, mark=o,mark options={scale=2.5}, blue]
  coordinates{
 (1, 0.0001)
(2, 0.0003619502416273894)
(3, 0.01161528927851426)
(4, 0.011240185961148862)
(5, 0.020629710124545945)
(6, 0.04456946331234274)
}; \addlegendentry{\huge LwF}

\addplot[smooth, mark=o,mark options={scale=2.5}, brown]
  coordinates{
(1, 0.0001)
(2, 0.0004347189318497285)
(3, 0.002763421749781449)
(4, 0.11480229862474794)
(5, 0.10332725954318246)
(6, 0.09466188072835648)
}; \addlegendentry{\huge iCaRL}

\end{axis}
\end{tikzpicture}{\label{subfig:bios-gap}}}
	\caption{{Evolution of target accuracy and $\mathrm{Gap}_{\mathbf g}^{\mathrm{RMS}}$ of models at different training stages. We observe that {\X} achieves a fine balance between accuracy and TPR-GAP.}}
	\label{fig:loss-evolution}
\end{figure}

We observe that {\X} (joint) is competitive with state-of-the-art debiasing frameworks AdS and FaRM. It is interesting to note that incrementally trained {\X} achieves better DP scores than jointly trained debiasing frameworks. 
We report the target accuracy and $\mathrm{Gap}_{\mathbf g}^{\mathrm{RMS}}$ metric across training stages. In Figure~\ref{fig:loss-evolution}(a), {\X} outperforms most baselines, marginally falling short of iCaRL in the final training stages. However, the $\mathrm{Gap}_{\mathbf g}^{\mathrm{RMS}}$ metric (in Figure~\ref{fig:loss-evolution}(b)) for {\X} is much better than iCaRL. LwF-based systems also achieve low scores but this is because of underfitting as evident from their low target accuracies. 

\subsection{Analysis}

In this section, we perform several analysis experiments to investigate the functioning of {\X}.

\noindent \textbf{Task ablations}.\label{sec:task-abl} We vary the number of classes that {\X} is presented with at a given training stage {and} report the average accuracy and fairness scores on Biographies dataset.
In Table~\ref{tab:step-ablation}, we observe a significant drop in target performance when the number of classes (in a training stage) are reduced 
accompanied by an improvement in DP, reflecting the tradeoff between fairness and utility as noted by \cite{zhao2019inherent}. 
The complete results for all sampling strategies are reported in Appendix~\ref{sec:num-tasks}.

\begin{table}[h!]
	\centering
	\resizebox{0.3\textwidth}{!}{
\begin{tabular}{c | c c c}
    \toprule[1pt]
      \# classes & Acc. ($\uparrow$) & DP ($\downarrow$) & $\mathrm{Gap}_{\mathbf g}^{\mathrm{RMS}}$ ($\downarrow$)\\ 
      \midrule[0.5pt]
      2 & 89.47 & 0.26 & 0.046 \\ 
      5 &  97.49 & 0.35 & 0.032\\ 
      10 &  98.57 & 0.39 & 0.031\\ 
    \bottomrule[1pt]
\end{tabular}
}
	\caption{{Performance of {\X} with varying number of classes per training stage. We observe an improved performance when the class count per training stage is increased.}}
	\label{tab:step-ablation}
\end{table}

\begin{figure}[h!]
	\centering
	\includegraphics[width=0.23\textwidth, keepaspectratio]{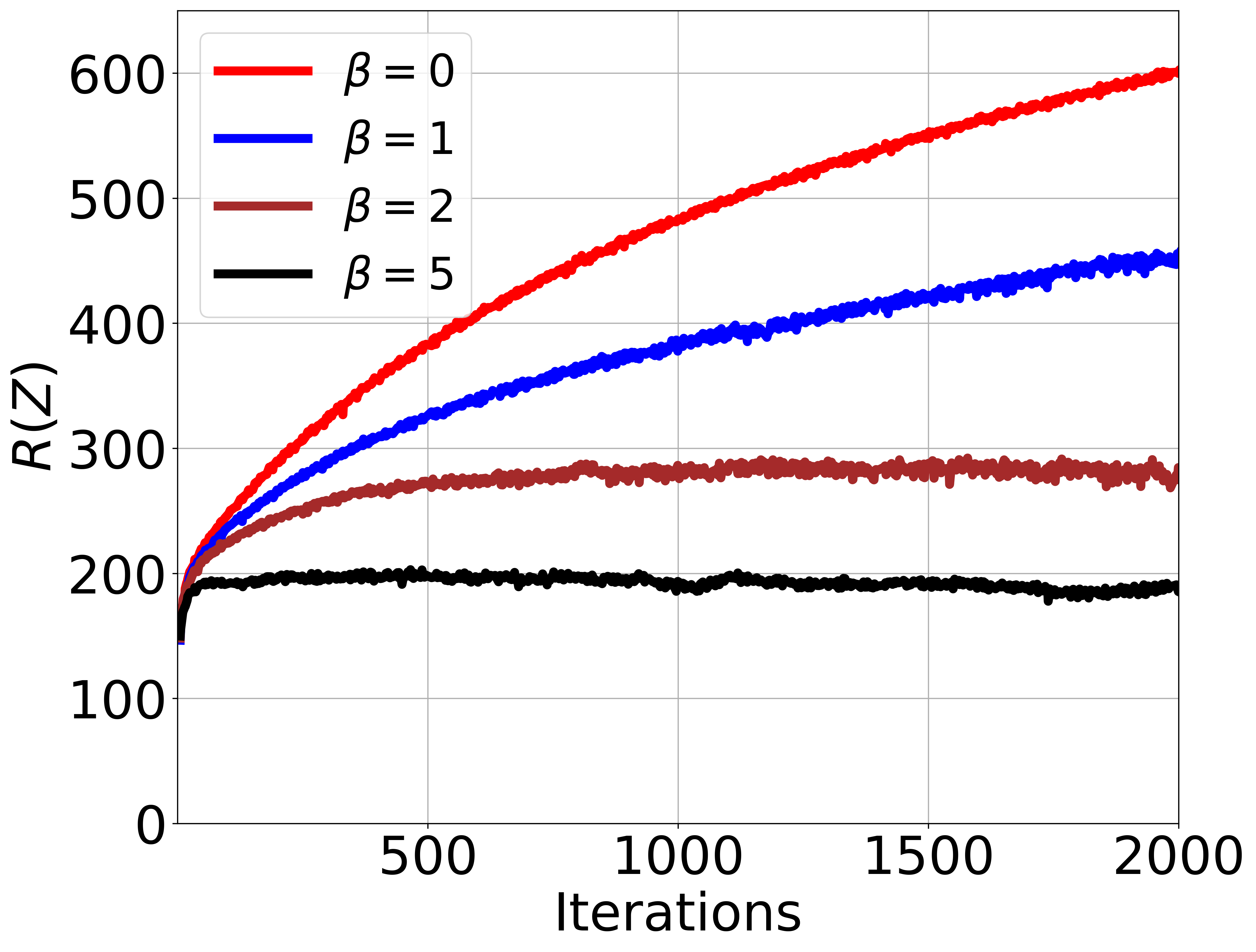}
	\includegraphics[width=0.22\textwidth, keepaspectratio]{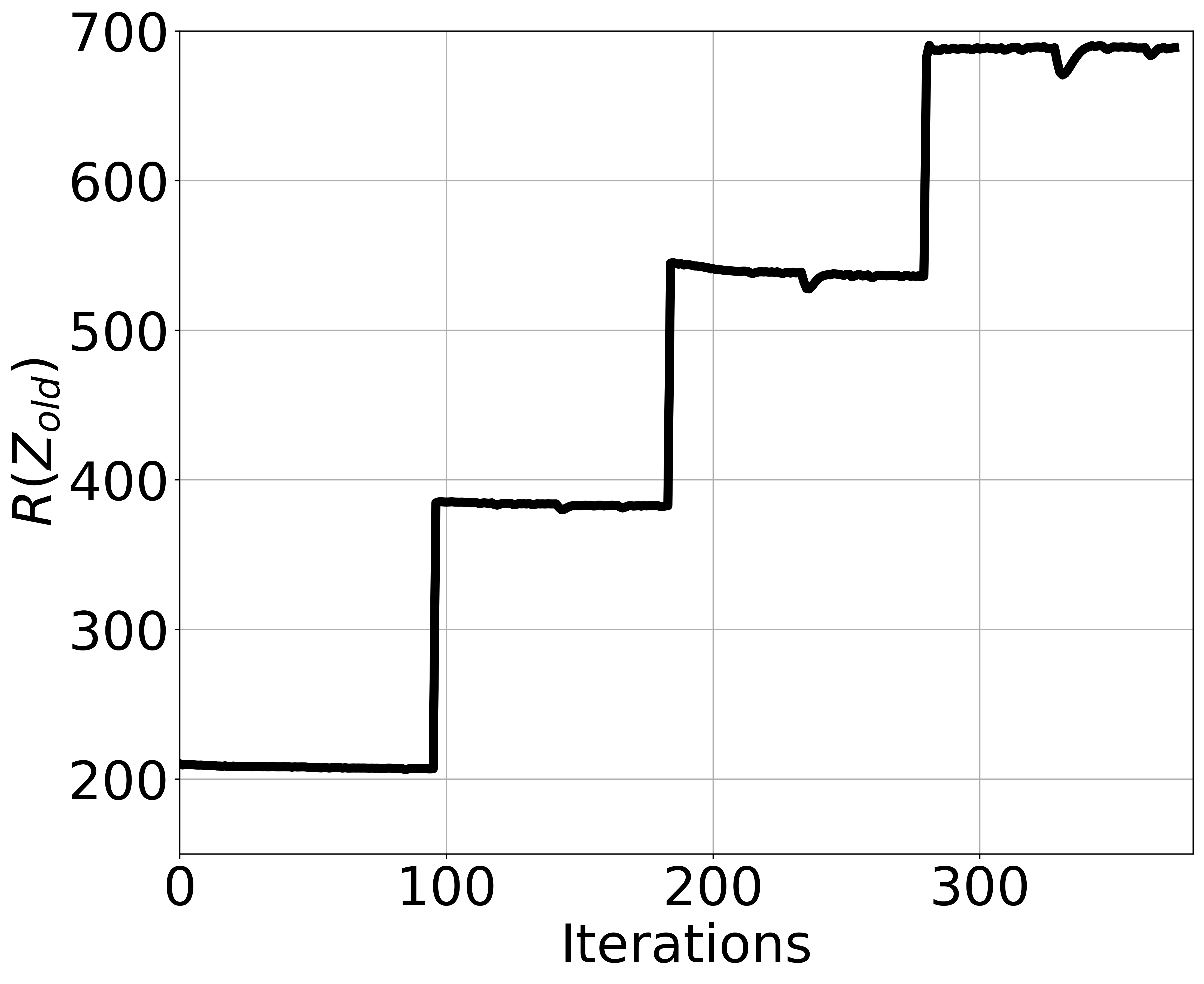}
	\caption{Evolution of rate-distortion function $R(Z)$ of representations during the training process of {\X}.}
	\label{fig:compact}
\end{figure}

\noindent \textbf{Evolution of feature space}.\label{sec:compact} We investigate the  evolution of $R(Z)$  in {\X}, which captures the volume of the feature space. In Figure~\ref{fig:compact}(a), we observe that having the debiasing component ($\beta>0$ in Equation~\ref{eqn:enc}) keeps the overall feature space compact (low $R(Z)$ values), which helps in incremental learning. However, a very high value of $\beta$ can prevent the feature space from expanding at all (shown in Figure~\ref{fig:compact}(a)) thereby resulting in poor target task performance.

In Figure~\ref{fig:compact}(b), we visualize the rate-distortion of the exemplar samples $R(Z_{old})$ as the training progressed. We observe a sharp rise in $R(Z_{old})$ at certain iterations, where new classes are added at every training stage. This shows that {\X} is able to leverage its compact feature space to gracefully accommodate representations from new tasks.

\begin{figure}[h!]
	\centering
	\includegraphics[width=0.17\textwidth, keepaspectratio]{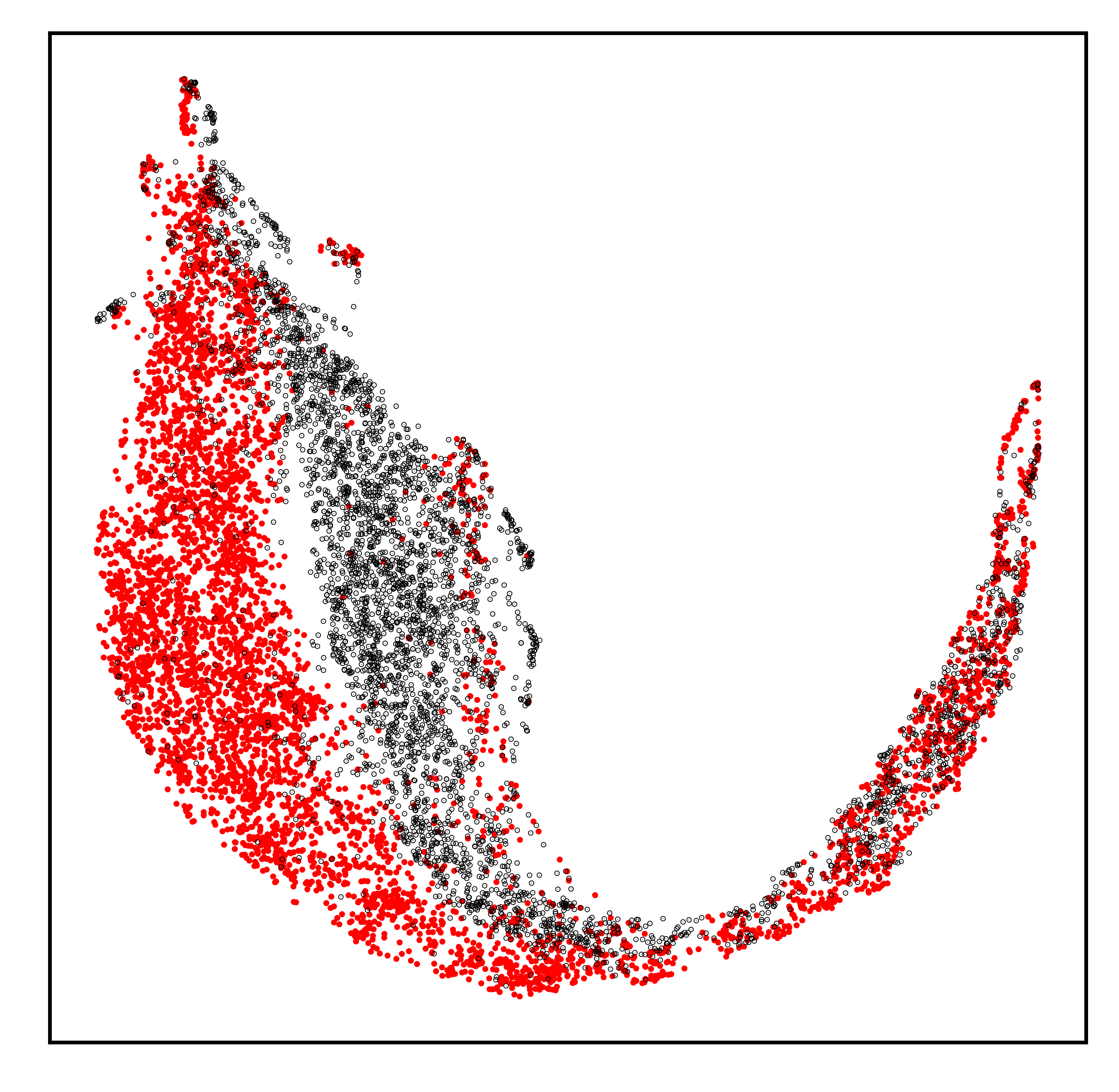}
	\includegraphics[width=0.17\textwidth, keepaspectratio]{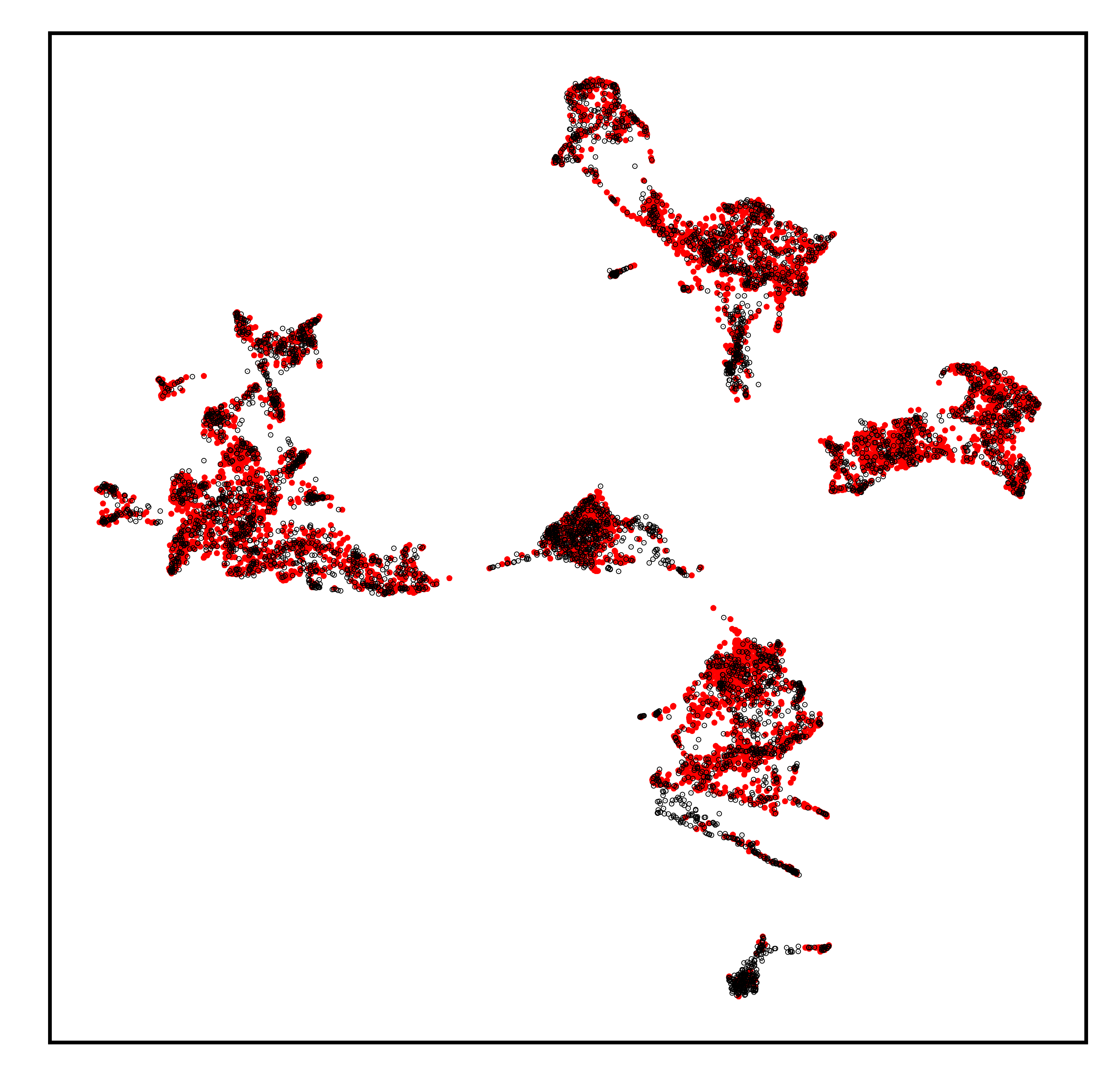}
	\caption{{UMAP projections of representations from {\X} before and after all the training stages.}}
	\label{fig:emb_vis}
\end{figure}

\noindent \textbf{Visualization}. We visualize the UMAP~\citep{mcinnes2018umap} feature projections before and after the debiasing process in Biographies dataset. The feature vectors are color-coded according to the protected attribute (gender). In Figure~\ref{fig:emb_vis}, we observe that before debiasing (left) it is easier to distinguish features, and after debiasing (right) features from both gender encompass similar subspaces.

We report additional analysis experiments to investigate the memory usage, sample efficiency, robustness, and effect of exemplar size on {\X}'s performance in Appendix~\ref{sec:addl-analys}.

\section{Conclusion}
In this work, we tackle the problem of learning fair representations in an incremental learning setting. To achieve this, we proposed \textbf{F}airness-\textbf{a}ware \textbf{I}ncremental \textbf{R}epresentation \textbf{L}earning ({\X}), a representation learning system that can make fair decisions while learning new tasks by controlling the rate-distortion function of representations. Empirical evaluations show that {\X} is able to make fair decisions outperforming prior baselines, even in scenarios where the target and protected attributes are strongly correlated. Through extensive analysis, we observe that the debiasing framework at the core of {\X} is able to keep the feature compact, which helps {\X} to learn new tasks in an incremental fashion. Our framework, {\X} can make fair decisions with incremental access to unseen tasks. Such systems will be crucial for achieving fairness in the wild, as learning systems are increasingly being deployed to critical applications. Future work can focus on developing incrementally trained fair decision-making systems with minimal reliance on protected attribute annotations.

\section*{Acknowledgement}
This research project was supported in part by Amazon Research Awards.

\bibliography{aaai23}

\begin{thebibliography}{47}
\providecommand{\natexlab}[1]{#1}

\bibitem[{Apfelbaum et~al.(2010)Apfelbaum, Pauker, Sommers, and
  Ambady}]{apfelbaum2010blind}
Apfelbaum, E.~P.; Pauker, K.; Sommers, S.~R.; and Ambady, N. 2010.
\newblock In blind pursuit of racial equality?
\newblock \emph{Psychological science}, 21(11): 1587--1592.

\bibitem[{Bahng et~al.(2020)Bahng, Chun, Yun, Choo, and Oh}]{bahng2020learning}
Bahng, H.; Chun, S.; Yun, S.; Choo, J.; and Oh, S.~J. 2020.
\newblock Learning De-biased Representations with Biased Representations.
\newblock In \emph{Proceedings of the 37th International Conference on Machine
  Learning, {ICML} 2020, 13-18 July 2020, Virtual Event}, volume 119 of
  \emph{Proceedings of Machine Learning Research}, 528--539. {PMLR}.

\bibitem[{Barrett et~al.(2019)Barrett, Kementchedjhieva, Elazar, Elliott, and
  S{\o}gaard}]{barrett-etal-2019-adversarial}
Barrett, M.; Kementchedjhieva, Y.; Elazar, Y.; Elliott, D.; and S{\o}gaard, A.
  2019.
\newblock Adversarial Removal of Demographic Attributes Revisited.
\newblock In \emph{Proceedings of the 2019 Conference on Empirical Methods in
  Natural Language Processing and the 9th International Joint Conference on
  Natural Language Processing (EMNLP-IJCNLP)}, 6330--6335. Hong Kong, China:
  Association for Computational Linguistics.

\bibitem[{Blodgett, Green, and O{'}Connor(2016)}]{blodgett2016demographic}
Blodgett, S.~L.; Green, L.; and O{'}Connor, B. 2016.
\newblock Demographic Dialectal Variation in Social Media: A Case Study of
  {A}frican-{A}merican {E}nglish.
\newblock In \emph{Proceedings of the 2016 Conference on Empirical Methods in
  Natural Language Processing}, 1119--1130. Austin, Texas: Association for
  Computational Linguistics.

\bibitem[{Bolukbasi et~al.(2016)Bolukbasi, Chang, Zou, Saligrama, and
  Kalai}]{bolukbasi2016man}
Bolukbasi, T.; Chang, K.; Zou, J.~Y.; Saligrama, V.; and Kalai, A.~T. 2016.
\newblock Man is to Computer Programmer as Woman is to Homemaker? Debiasing
  Word Embeddings.
\newblock In Lee, D.~D.; Sugiyama, M.; von Luxburg, U.; Guyon, I.; and Garnett,
  R., eds., \emph{Advances in Neural Information Processing Systems 29: Annual
  Conference on Neural Information Processing Systems 2016, December 5-10,
  2016, Barcelona, Spain}, 4349--4357.

\bibitem[{Castro et~al.(2018)Castro, Mar{\'\i}n-Jim{\'e}nez, Guil, Schmid, and
  Alahari}]{castro2018end}
Castro, F.~M.; Mar{\'\i}n-Jim{\'e}nez, M.~J.; Guil, N.; Schmid, C.; and
  Alahari, K. 2018.
\newblock End-to-end incremental learning.
\newblock In \emph{Proceedings of the European conference on computer vision
  (ECCV)}, 233--248.

\bibitem[{Chan et~al.(2021)Chan, Yu, You, Qi, Wright, and Ma}]{chan2021redunet}
Chan, K. H.~R.; Yu, Y.; You, C.; Qi, H.; Wright, J.; and Ma, Y. 2021.
\newblock ReduNet: A white-box deep network from the principle of maximizing
  rate reduction.
\newblock \emph{ArXiv preprint}, abs/2105.10446.

\bibitem[{Chaudhry et~al.(2019{\natexlab{a}})Chaudhry, Ranzato, Rohrbach, and
  Elhoseiny}]{chaudhry2018efficient}
Chaudhry, A.; Ranzato, M.; Rohrbach, M.; and Elhoseiny, M. 2019{\natexlab{a}}.
\newblock Efficient Lifelong Learning with {A-GEM}.
\newblock In \emph{7th International Conference on Learning Representations,
  {ICLR} 2019, New Orleans, LA, USA, May 6-9, 2019}. OpenReview.net.

\bibitem[{Chaudhry et~al.(2019{\natexlab{b}})Chaudhry, Rohrbach, Elhoseiny,
  Ajanthan, Dokania, Torr, and Ranzato}]{chaudhry2019tiny}
Chaudhry, A.; Rohrbach, M.; Elhoseiny, M.; Ajanthan, T.; Dokania, P.~K.; Torr,
  P.~H.; and Ranzato, M. 2019{\natexlab{b}}.
\newblock On tiny episodic memories in continual learning.
\newblock \emph{ArXiv preprint}, abs/1902.10486.

\bibitem[{Cheng et~al.(2020)Cheng, Hao, Yuan, Si, and Carin}]{fairfil}
Cheng, P.; Hao, W.; Yuan, S.; Si, S.; and Carin, L. 2020.
\newblock FairFil: Contrastive Neural Debiasing Method for Pretrained Text
  Encoders.
\newblock In \emph{International Conference on Learning Representations}.

\bibitem[{Chowdhury and Chaturvedi(2022)}]{chowdhury2022learning}
Chowdhury, S. B.~R.; and Chaturvedi, S. 2022.
\newblock {Learning Fair Representations via Rate-Distortion Maximization}.
\newblock \emph{Transactions of the Association for Computational Linguistics},
  10: 1159--1174.

\bibitem[{Chowdhury et~al.(2021)Chowdhury, Ghosh, Li, Oliva, Srivastava, and
  Chaturvedi}]{basu-roy-chowdhury-etal-2021-adversarial}
Chowdhury, S. B.~R.; Ghosh, S.; Li, Y.; Oliva, J.; Srivastava, S.; and
  Chaturvedi, S. 2021.
\newblock Adversarial Scrubbing of Demographic Information for Text
  Classification.
\newblock In \emph{Proceedings of the 2021 Conference on Empirical Methods in
  Natural Language Processing}, 550--562. Online and Punta Cana, Dominican
  Republic: Association for Computational Linguistics.

\bibitem[{Cover(1999)}]{cover1999elements}
Cover, T.~M. 1999.
\newblock \emph{Elements of information theory}.
\newblock John Wiley \& Sons.

\bibitem[{Dastin(2018)}]{dastin2018amazon}
Dastin, J. 2018.
\newblock Amazon scraps secret AI recruiting tool that showed bias against
  women.
\newblock In \emph{Ethics of Data and Analytics}, 296--299. Auerbach
  Publications.

\bibitem[{De-Arteaga et~al.(2019)De-Arteaga, Romanov, Wallach, Chayes, Borgs,
  Chouldechova, Geyik, Kenthapadi, and Kalai}]{de2019bias}
De-Arteaga, M.; Romanov, A.; Wallach, H.; Chayes, J.; Borgs, C.; Chouldechova,
  A.; Geyik, S.; Kenthapadi, K.; and Kalai, A.~T. 2019.
\newblock Bias in bios: A case study of semantic representation bias in a
  high-stakes setting.
\newblock In \emph{proceedings of the Conference on Fairness, Accountability,
  and Transparency}, 120--128.

\bibitem[{Dixon et~al.(2018)Dixon, Li, Sorensen, Thain, and
  Vasserman}]{dixon2018measuring}
Dixon, L.; Li, J.; Sorensen, J.; Thain, N.; and Vasserman, L. 2018.
\newblock Measuring and mitigating unintended bias in text classification.
\newblock In \emph{Proceedings of the 2018 AAAI/ACM Conference on AI, Ethics,
  and Society}, 67--73.

\bibitem[{Elazar and Goldberg(2018)}]{elazar2018adversarial}
Elazar, Y.; and Goldberg, Y. 2018.
\newblock Adversarial Removal of Demographic Attributes from Text Data.
\newblock In \emph{Proceedings of the 2018 Conference on Empirical Methods in
  Natural Language Processing}, 11--21. Brussels, Belgium: Association for
  Computational Linguistics.

\bibitem[{Elazar et~al.(2021)Elazar, Ravfogel, Jacovi, and
  Goldberg}]{elazar2021amnesic}
Elazar, Y.; Ravfogel, S.; Jacovi, A.; and Goldberg, Y. 2021.
\newblock Amnesic probing: Behavioral explanation with amnesic counterfactuals.
\newblock \emph{Transactions of the Association for Computational Linguistics},
  9: 160--175.

\bibitem[{Frieze(1974)}]{frieze1974cost}
Frieze, A.~M. 1974.
\newblock A cost function property for plant location problems.
\newblock \emph{Mathematical Programming}, 7(1): 245--248.

\bibitem[{Goodfellow et~al.(2014)Goodfellow, Pouget-Abadie, Mirza, Xu,
  Warde-Farley, Ozair, Courville, and Bengio}]{goodfellow2014generative}
Goodfellow, I.~J.; Pouget-Abadie, J.; Mirza, M.; Xu, B.; Warde-Farley, D.;
  Ozair, S.; Courville, A.; and Bengio, Y. 2014.
\newblock Generative adversarial networks.
\newblock \emph{ArXiv preprint}, abs/1406.2661.

\bibitem[{Kirkpatrick et~al.(2017)Kirkpatrick, Pascanu, Rabinowitz, Veness,
  Desjardins, Rusu, Milan, Quan, Ramalho, Grabska-Barwinska
  et~al.}]{kirkpatrick2017overcoming}
Kirkpatrick, J.; Pascanu, R.; Rabinowitz, N.; Veness, J.; Desjardins, G.; Rusu,
  A.~A.; Milan, K.; Quan, J.; Ramalho, T.; Grabska-Barwinska, A.; et~al. 2017.
\newblock Overcoming catastrophic forgetting in neural networks.
\newblock \emph{Proceedings of the national academy of sciences}, 114(13):
  3521--3526.

\bibitem[{Krause and Golovin(2014)}]{krause2014submodular}
Krause, A.; and Golovin, D. 2014.
\newblock Submodular function maximization.
\newblock \emph{Tractability}, 3: 71--104.

\bibitem[{Larson et~al.(2016)Larson, Mattu, Kirchner, and
  Angwin}]{larson2016we}
Larson, J.; Mattu, S.; Kirchner, L.; and Angwin, J. 2016.
\newblock How we analyzed the COMPAS recidivism algorithm.
\newblock \emph{ProPublica (5 2016)}, 9(1): 3--3.

\bibitem[{LeCun et~al.(1998)LeCun, Bottou, Bengio, and
  Haffner}]{lecun1998gradient}
LeCun, Y.; Bottou, L.; Bengio, Y.; and Haffner, P. 1998.
\newblock Gradient-based learning applied to document recognition.
\newblock \emph{Proceedings of the IEEE}, 86(11): 2278--2324.

\bibitem[{Li et~al.(2019)Li, Zhou, Wu, Socher, and Xiong}]{li2019learn}
Li, X.; Zhou, Y.; Wu, T.; Socher, R.; and Xiong, C. 2019.
\newblock Learn to Grow: {A} Continual Structure Learning Framework for
  Overcoming Catastrophic Forgetting.
\newblock In Chaudhuri, K.; and Salakhutdinov, R., eds., \emph{Proceedings of
  the 36th International Conference on Machine Learning, {ICML} 2019, 9-15 June
  2019, Long Beach, California, {USA}}, volume~97 of \emph{Proceedings of
  Machine Learning Research}, 3925--3934. {PMLR}.

\bibitem[{Li, Baldwin, and Cohn(2018)}]{li2018towards}
Li, Y.; Baldwin, T.; and Cohn, T. 2018.
\newblock Towards Robust and Privacy-preserving Text Representations.
\newblock In \emph{Proceedings of the 56th Annual Meeting of the Association
  for Computational Linguistics (Volume 2: Short Papers)}, 25--30. Melbourne,
  Australia: Association for Computational Linguistics.

\bibitem[{Li and Hoiem(2017)}]{li2017learning}
Li, Z.; and Hoiem, D. 2017.
\newblock Learning without forgetting.
\newblock \emph{IEEE transactions on pattern analysis and machine
  intelligence}, 40(12): 2935--2947.

\bibitem[{Long et~al.(2015)Long, Cao, Wang, and Jordan}]{long2015learning}
Long, M.; Cao, Y.; Wang, J.; and Jordan, M.~I. 2015.
\newblock Learning Transferable Features with Deep Adaptation Networks.
\newblock In Bach, F.~R.; and Blei, D.~M., eds., \emph{Proceedings of the 32nd
  International Conference on Machine Learning, {ICML} 2015, Lille, France,
  6-11 July 2015}, volume~37 of \emph{{JMLR} Workshop and Conference
  Proceedings}, 97--105. JMLR.org.

\bibitem[{Ma et~al.(2007)Ma, Derksen, Hong, and Wright}]{ma2007segmentation}
Ma, Y.; Derksen, H.; Hong, W.; and Wright, J. 2007.
\newblock Segmentation of multivariate mixed data via lossy data coding and
  compression.
\newblock \emph{IEEE transactions on pattern analysis and machine
  intelligence}, 29(9): 1546--1562.

\bibitem[{McCloskey and Cohen(1989)}]{mccloskey1989catastrophic}
McCloskey, M.; and Cohen, N.~J. 1989.
\newblock Catastrophic interference in connectionist networks: The sequential
  learning problem.
\newblock In \emph{Psychology of learning and motivation}, volume~24, 109--165.
  Elsevier.

\bibitem[{McInnes et~al.(2018)McInnes, Healy, Saul, and
  Großberger}]{mcinnes2018umap}
McInnes, L.; Healy, J.; Saul, N.; and Großberger, L. 2018.
\newblock UMAP: Uniform Manifold Approximation and Projection.
\newblock \emph{Journal of Open Source Software}, 3(29): 861.

\bibitem[{Mehrabi et~al.(2021)Mehrabi, Morstatter, Saxena, Lerman, and
  Galstyan}]{mehrabi2019survey}
Mehrabi, N.; Morstatter, F.; Saxena, N.; Lerman, K.; and Galstyan, A. 2021.
\newblock A survey on bias and fairness in machine learning.
\newblock \emph{ACM Computing Surveys (CSUR)}, 54(6): 1--35.

\bibitem[{Ravfogel et~al.(2020)Ravfogel, Elazar, Gonen, Twiton, and
  Goldberg}]{ravfogel2020null}
Ravfogel, S.; Elazar, Y.; Gonen, H.; Twiton, M.; and Goldberg, Y. 2020.
\newblock Null It Out: Guarding Protected Attributes by Iterative Nullspace
  Projection.
\newblock In \emph{Proceedings of the 58th Annual Meeting of the Association
  for Computational Linguistics}, 7237--7256. Online: Association for
  Computational Linguistics.

\bibitem[{Rebuffi et~al.(2017)Rebuffi, Kolesnikov, Sperl, and
  Lampert}]{rebuffi2017icarl}
Rebuffi, S.; Kolesnikov, A.; Sperl, G.; and Lampert, C.~H. 2017.
\newblock iCaRL: Incremental Classifier and Representation Learning.
\newblock In \emph{2017 {IEEE} Conference on Computer Vision and Pattern
  Recognition, {CVPR} 2017, Honolulu, HI, USA, July 21-26, 2017}, 5533--5542.
  {IEEE} Computer Society.

\bibitem[{Rezaei et~al.(2021)Rezaei, Liu, Memarrast, and
  Ziebart}]{rezaei2021robust}
Rezaei, A.; Liu, A.; Memarrast, O.; and Ziebart, B.~D. 2021.
\newblock Robust fairness under covariate shift.
\newblock In \emph{Proceedings of the AAAI Conference on Artificial
  Intelligence}, volume~35, 9419--9427.

\bibitem[{Romanov et~al.(2019)Romanov, De-Arteaga, Wallach, Chayes, Borgs,
  Chouldechova, Geyik, Kenthapadi, Rumshisky, and Kalai}]{biasbios2}
Romanov, A.; De-Arteaga, M.; Wallach, H.; Chayes, J.; Borgs, C.; Chouldechova,
  A.; Geyik, S.; Kenthapadi, K.; Rumshisky, A.; and Kalai, A. 2019.
\newblock What{'}s in a Name? {R}educing Bias in Bios without Access to
  Protected Attributes.
\newblock In \emph{Proceedings of the 2019 Conference of the North {A}merican
  Chapter of the Association for Computational Linguistics: Human Language
  Technologies, Volume 1 (Long and Short Papers)}, 4187--4195. Minneapolis,
  Minnesota: Association for Computational Linguistics.

\bibitem[{Rusu et~al.(2016)Rusu, Rabinowitz, Desjardins, Soyer, Kirkpatrick,
  Kavukcuoglu, Pascanu, and Hadsell}]{rusu2016progressive}
Rusu, A.~A.; Rabinowitz, N.~C.; Desjardins, G.; Soyer, H.; Kirkpatrick, J.;
  Kavukcuoglu, K.; Pascanu, R.; and Hadsell, R. 2016.
\newblock Progressive neural networks.
\newblock \emph{ArXiv preprint}, abs/1606.04671.

\bibitem[{Shah, Schwartz, and Hovy(2020)}]{shah2019predictive}
Shah, D.~S.; Schwartz, H.~A.; and Hovy, D. 2020.
\newblock Predictive Biases in Natural Language Processing Models: A Conceptual
  Framework and Overview.
\newblock In \emph{Proceedings of the 58th Annual Meeting of the Association
  for Computational Linguistics}, 5248--5264. Online: Association for
  Computational Linguistics.

\bibitem[{Singh et~al.(2021)Singh, Singh, Mhasawade, and
  Chunara}]{singh2021fairness}
Singh, H.; Singh, R.; Mhasawade, V.; and Chunara, R. 2021.
\newblock Fairness violations and mitigation under covariate shift.
\newblock In \emph{Proceedings of the 2021 ACM Conference on Fairness,
  Accountability, and Transparency}, 3--13.

\bibitem[{Tong et~al.(2022)Tong, Dai, Wu, Li, Yi, and Ma}]{tong2022incremental}
Tong, S.; Dai, X.; Wu, Z.; Li, M.; Yi, B.; and Ma, Y. 2022.
\newblock Incremental Learning of Structured Memory via Closed-Loop
  Transcription.
\newblock \emph{ArXiv preprint}, abs/2202.05411.

\bibitem[{Yu et~al.(2020)Yu, Chan, You, Song, and Ma}]{yu2020learning}
Yu, Y.; Chan, K. H.~R.; You, C.; Song, C.; and Ma, Y. 2020.
\newblock Learning diverse and discriminative representations via the principle
  of maximal coding rate reduction.
\newblock \emph{ArXiv preprint}, abs/2006.08558.

\bibitem[{Zemel et~al.(2013)Zemel, Wu, Swersky, Pitassi, and
  Dwork}]{zemel2013learning}
Zemel, R.~S.; Wu, Y.; Swersky, K.; Pitassi, T.; and Dwork, C. 2013.
\newblock Learning Fair Representations.
\newblock In \emph{Proceedings of the 30th International Conference on Machine
  Learning, {ICML} 2013, Atlanta, GA, USA, 16-21 June 2013}, volume~28 of
  \emph{{JMLR} Workshop and Conference Proceedings}, 325--333. JMLR.org.

\bibitem[{Zenke, Poole, and Ganguli(2017)}]{zenke2017continual}
Zenke, F.; Poole, B.; and Ganguli, S. 2017.
\newblock Continual Learning Through Synaptic Intelligence.
\newblock In Precup, D.; and Teh, Y.~W., eds., \emph{Proceedings of the 34th
  International Conference on Machine Learning, {ICML} 2017, Sydney, NSW,
  Australia, 6-11 August 2017}, volume~70 of \emph{Proceedings of Machine
  Learning Research}, 3987--3995. {PMLR}.

\bibitem[{Zhang, Lemoine, and Mitchell(2018)}]{zhang2018mitigating}
Zhang, B.~H.; Lemoine, B.; and Mitchell, M. 2018.
\newblock Mitigating unwanted biases with adversarial learning.
\newblock In \emph{Proceedings of the 2018 AAAI/ACM Conference on AI, Ethics,
  and Society}, 335--340.

\bibitem[{Zhang et~al.(2021)Zhang, Bifet, Zhang, Weiss, and
  Nejdl}]{zhang2021farf}
Zhang, W.; Bifet, A.; Zhang, X.; Weiss, J.~C.; and Nejdl, W. 2021.
\newblock Farf: A fair and adaptive random forests classifier.
\newblock In \emph{Pacific-Asia Conference on Knowledge Discovery and Data
  Mining}, 245--256. Springer.

\bibitem[{Zhang and Ntoutsi(2019)}]{zhang2019faht}
Zhang, W.; and Ntoutsi, E. 2019.
\newblock FAHT: An Adaptive Fairness-aware Decision Tree Classifier.
\newblock In \emph{IJCAI}.

\bibitem[{Zhao and Gordon(2019)}]{zhao2019inherent}
Zhao, H.; and Gordon, G.~J. 2019.
\newblock Inherent Tradeoffs in Learning Fair Representations.
\newblock In Wallach, H.~M.; Larochelle, H.; Beygelzimer, A.;
  d'Alch{\'{e}}{-}Buc, F.; Fox, E.~B.; and Garnett, R., eds., \emph{Advances in
  Neural Information Processing Systems 32: Annual Conference on Neural
  Information Processing Systems 2019, NeurIPS 2019, December 8-14, 2019,
  Vancouver, BC, Canada}, 15649--15659.

\end{thebibliography}

\clearpage









\appendix
\section{Appendix}

\section{Method Implementation}

\noindent \textbf{MCR\textsuperscript{2} Illustration}.
\label{sec:mcr-ill}
 We illustrate the implementation of MCR\textsuperscript{2} for multi-label classification using a simple example. Specifically, we focus on the construction of the global membership matrix $\Pi$, which is a collection of diagonal matrices $\Pi = \{\Pi_j\}_{j=1}^k$ for each of the $k$ classes. For each class, $\Pi_j$ is defined as:

\begin{equation}
    \Pi_j = 
    \begin{bmatrix}
    \pi_{1j} & 0 & \dots & 0 \\
    0 & \pi_{2j} & \dots & 0 \\
    \vdots & \vdots & \ddots & \vdots \\
    0 & 0 & \dots & \pi_{nj}
  \end{bmatrix} \in \mathbb{R}^{n \times n}
\end{equation}

where $n$ is the number of data samples. For multi-label classification, where a data sample can be a member of a single class $\pi_{ij} \in \{0, 1\}$. Let us consider a dataset with 4 samples $Z = \{z_i\}_{i=1}^4$ from 2 distinct classes with labels $y = \{1, 2, 1, 1\}$. The corresponding membership matrices are:
\begin{equation}
    \Pi_1 = 
    \begin{bmatrix}
    1 & 0 & 0 & 0 \\
    0 & 0 & 0 & 0 \\
    0 & 0 & 1 & 0 \\
    0 & 0 & 0 & 1
  \end{bmatrix}, \;\;
    \Pi_2 = 
    \begin{bmatrix}
    0 & 0 & 0 & 0 \\
    0 & 1 & 0 & 0 \\
    0 & 0 & 0 & 0 \\
    0 & 0 & 0 & 0
  \end{bmatrix}
\end{equation}

$\Pi = \{\Pi_1, \Pi_2\}$ matrices are used for computing the second term in MCR\textsuperscript{2} ($R_c(Z, \epsilon|\Pi)$ in Equation~\ref{eqn:mcr}). This process is followed to encode prediction label $y \in \mathbb{R}^{n}$ information  for multi-label classification

\begin{figure}[h!]
    \centering
    \includegraphics[width=0.5\textwidth, keepaspectratio]{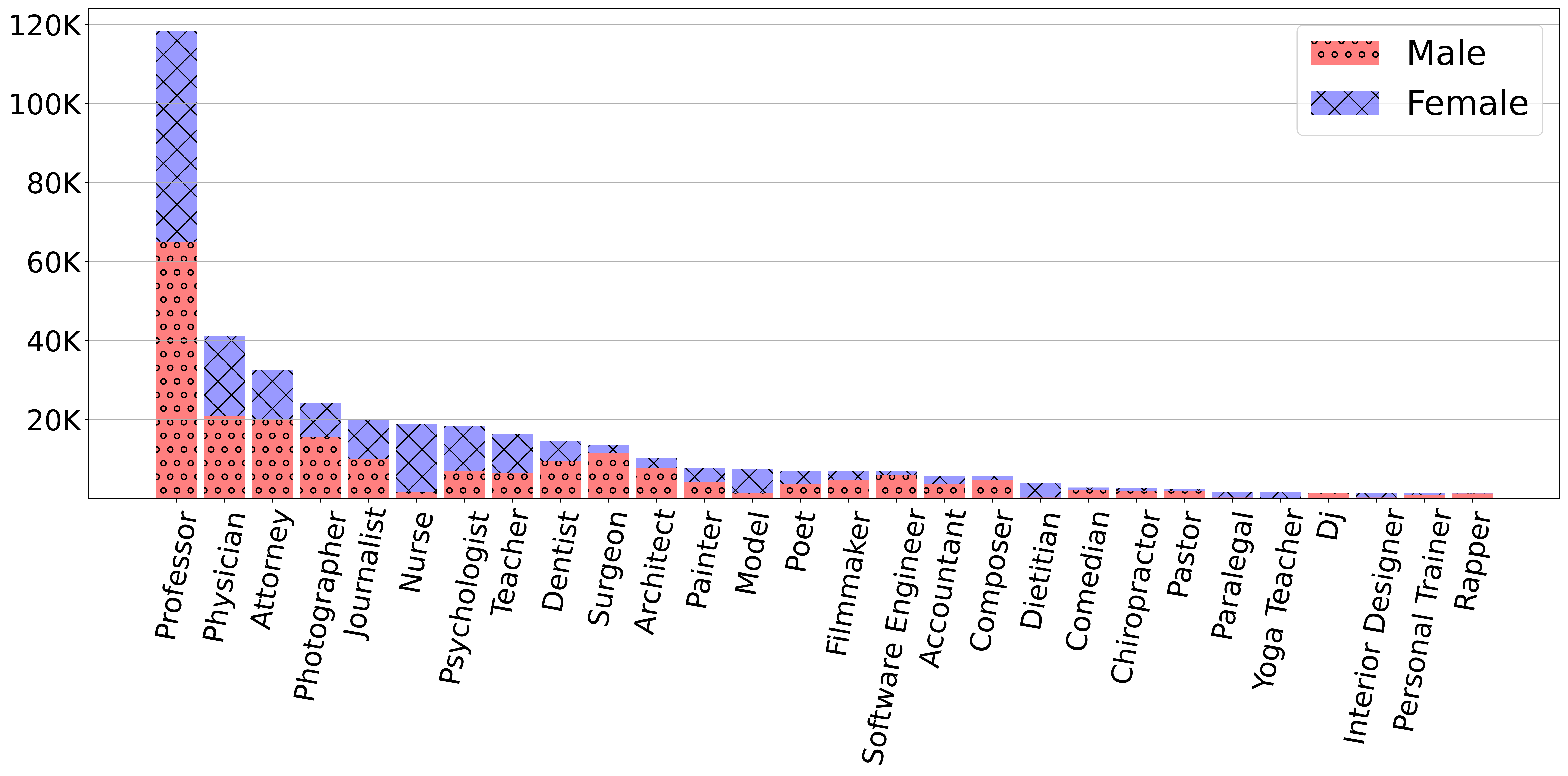}
    \caption{{Demographic distribution of different professions in Biographies dataset.}}
    \label{fig:bios-demog.}
\end{figure}

\noindent\textbf{Dataset Statistics}.
\label{sec:data-stats} We repurpose the original MNIST dataset to generate Biased MNIST. It has 60K training samples, and 10K test samples. The details of the coloring process is discussed in Section~\ref{sec:dataset}.  Biographies dataset has 257K training samples, 40K dev samples, and 99K test samples. Every biography in this dataset is associated with one of 28 profession categories, and a binary gender label. Biographies contain personally identifiable information that are already available on the internet. \citet{de2019bias} curated the dataset with appropriate approval and publicly released it for academic usage. We did not perform any additional data collection. The demographic distribution of Biographies dataset is shown in Figure~\ref{fig:bios-demog.}. We observe that the gender distribution is skewed for certain professions like {surgeon}, {dietician} etc.

\noindent\textbf{Experimental Setup}.
\label{sec:impl} For Biased MNIST dataset, our feature encoder $\phi$ is a two-layer CNN network with kernel size $5 \times 5$, followed by a fully-connected network. For Biographies, $\phi$ is a BERT\textsubscript{\textit{base}} encoder followed by a 2-layer MLP. The discriminator is a 2-layer fully connected network. We use the same feature encoder for all setups including the baselines. The model was trained using Adam Optimizer with a learning rate of $2\times 10^{-5}$. All experiments were implemented using PyTorch~\citep{NEURIPS2019_9015}. We trained our model for 2 epochs at each training stages using a single GeForce GTX 2080 Ti GPU. We set $\beta=\gamma=\eta=1$ for Biographies, and $\beta=\eta=0.01, \gamma=1$ for Biased MNIST dataset respectively.  All hyperparameters were tuned on the development set of each dataset. Implementation of the baseline approach iCaRL was adapted from the open-sourced project.\footnote{https://github.com/donlee90/icarl} In Biased MNIST dataset, most digit classes have similar number of samples, therefore we present incremental learning systems classes in a random order. In Biographies dataset, we present incremental learning systems with classes in descending order of their size (number of samples in their class). The order of class sequence for Biographies is shown in Figure~\ref{fig:bios-demog.}. We provide the system with 5 classes at each training stage (except in the last trainng stage, where it has access to 3 classes). For both datasets, we select 20 samples per class as the exemplar set. This is quite small compared to the average class sizes in the training set: Biographies $\sim9.2$K, Biased MNIST $\sim6$K.

\section{Additional Analysis}
\label{sec:addl-analys}
\begin{figure}[h!]
    \centering
    
    \includegraphics[width=0.32\textwidth, keepaspectratio]{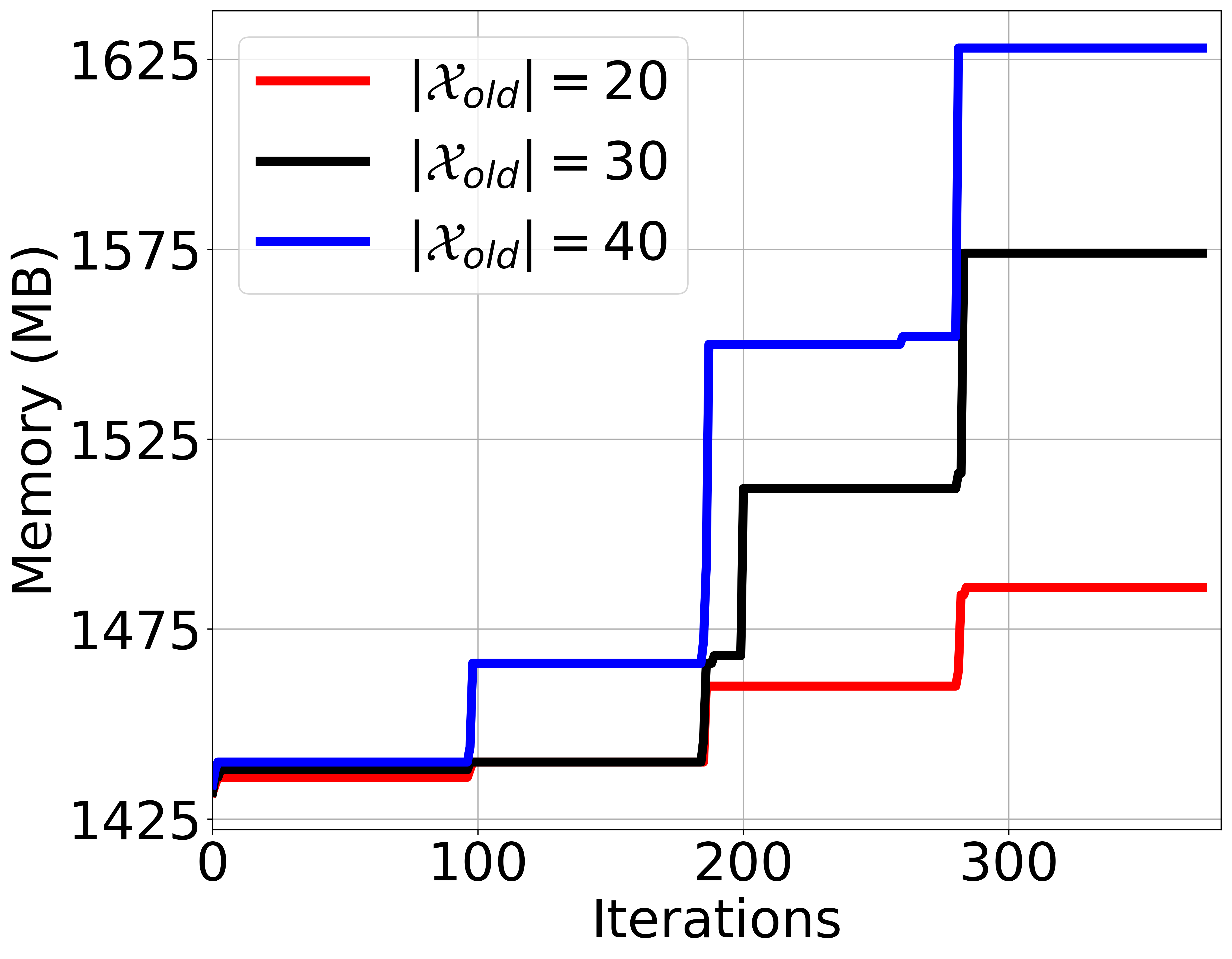}
    \caption{GPU memory usage over training iterations.}
    \label{fig:memory-evolution}
\end{figure}

\noindent \textbf{Memory Usage}. In this experiment, we report how the GPU memory usage varies during training {\X}. We perform this experiment on Biased MNIST dataset with random sampling strategy and varying exemplar set sizes. In Figure~\ref{fig:memory-evolution}, we observe that memory usage increases over training iterations as exemplar set becomes larger. However, the gain in memory usage at a training step is small ($<10\%$) compared to the total memory being used. We also observe that the growth in memory usage rises progressively as the exemplar set size increases.

\noindent \textbf{Ablation with number of tasks}.\label{sec:num-tasks} We extend the ablation experiments on the number of classes {\X} receives at each training stage for all sampling strategies.  In Table~\ref{tab:step-ablation-2}, we report the average performance metrics of {\X} on Biographies dataset using varying sampling strategies. Similar trends noted in Section~\ref{sec:task-abl} hold true for all settings. It is interesting to note that when the number of classes=2, prototypical sampling outperforms other strategies by a significant margin in terms of target accuracy.
\begin{table}[h!]
    \centering
    \resizebox{0.43\textwidth}{!}{
\begin{tabular}{l |c| c c c}
    \toprule[1pt]
      	Sampling & {\# classes} & Acc. ($\uparrow$) & DP ($\downarrow$) & $\mathrm{Gap}_{\mathbf g}^{\mathrm{RMS}}$ ($\downarrow$)\TBstrut\\ 
      \midrule[0.5pt]
       \multirow{3}{*}{\textit{random}} & 2 & 89.47 & \textbf{0.26} & 0.046 \Tstrut\\ 
      & 5 &  97.49 & 0.35 & 0.032\\ 
     & 10 &  \textbf{98.57} & 0.39 & \textbf{0.031}\Bstrut\\ 
     \midrule[0.5pt]
      \multirow{3}{*}{\textit{prototypical}} & 2 & 92.12 & \textbf{0.28} & 0.042 \Tstrut\\ 
      & 5 &  93.93 & 0.34 & 0.031\\ 
     & 10 &  \textbf{98.33} & 0.39 & \textbf{0.030}\Bstrut\\ 
     \midrule[0.5pt]
      \multirow{3}{*}{\textit{submodular}} & 2 & 86.29 & \textbf{0.24} & 0.033 \Tstrut\\ 
      & 5 &  94.37 & 0.35 & \textbf{0.022}\\ 
     & 10 &  \textbf{98.48} & 0.39 & 0.040\Bstrut\\ 
    \bottomrule[1pt]
\end{tabular}
}
    \caption{Performance of {\X} with varying number of classes on Biographies dataset.}
    \label{tab:step-ablation-2}
\end{table}

\noindent \textbf{Effect of the Debiasing Component}. In this experiment, we investigate the effect of the debiasing component on incremental learning task. We perform ablations of {\X} by changing the hyperparameters $\beta, \eta$ (Equation~\ref{eqn:enc-inc}). In Table~\ref{tab:debias-ablations}, we report the amortized accuracy, leakage and fairness metrics achieved by {\X} in different configurations on Biographies dataset. We observe that removing debiasing component completely (configuration: $\beta=0, \eta=0$), results in drop in target accuracy as well as leakage of protected attribute, resulting in poor fairness metric scores. We believe the poor performance in target task occurs as the feature space doesn't remain compact due to the removal of debiasing component, which hinders incremental learning. We observe similar results with ($\beta=0, \eta=1$) configuration, but the accuracy is slightly better due to the presence of debiasing component of $\mathcal{X}_{old}$, as it also helps in keeping the feature space compact.  Only having the debiasing component for $\mathcal{X}_{new}$ (configuration $\beta=1, \eta=0$),  results in better performance both in terms of accuracy and fairness metrics, falling slightly short of the configuration $(\beta=1, \eta=1)$.

\begin{table}[h!]
    \centering
    \resizebox{0.47\textwidth}{!}{
\begin{tabular}{c| c c c c}
    \toprule[1pt]
      	{Configuration} & Acc. ($\uparrow$)  & Leakage ($\downarrow$) & DP ($\downarrow$) & $\mathrm{Gap}_{\mathbf g}^{\mathrm{RMS}}$ ($\downarrow$)\TBstrut\\ 
      \midrule[0.5pt]
      ($\beta=0, \eta=0$) & 91.19 & 43.28 & 0.36 & 0.129 \\
      ($\beta=0, \eta=1$) & 94.78 & 41.97 & 0.37 & 0.094\\
      ($\beta=1, \eta=0$) &  95.92 & 11.33 & \textbf{0.34} & 0.034\\
      ($\beta=1, \eta=1$) &  \textbf{97.49} & \textbf{9.49} & 0.35 & \textbf{0.032}\\
    \bottomrule[1pt]
\end{tabular}
}
    \caption{Performance of {\X} on Biographies dataset in different configurations involving the debiasing component. The best results are highlighted in \textbf{bold}.}
    \label{tab:debias-ablations}
\end{table}

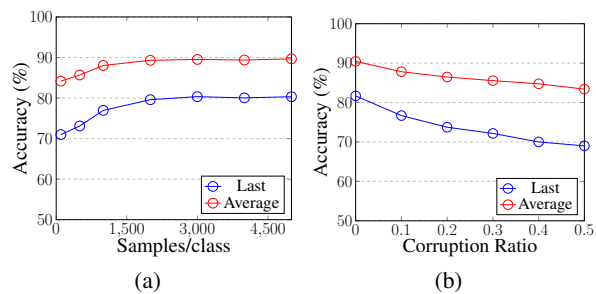
\begin{figure}[h!]
    \centering
    \subfloat[][\footnotesize ]{
    \resizebox{0.22\textwidth}{!}{
        \begin{tikzpicture}
\pgfplotsset{
    scale only axis,
    tick label style={font=\huge},
}

\begin{axis}[
  ymin=50, ymax=100,
  xmin=0, xmax=5000,
  xtick={0, 1500, 3000, 4500},
    ymajorgrids=true,
    grid style=dashed,
  xlabel={\Huge Samples/class},
  ylabel={\Huge Accuracy (\%)},
  legend pos=south east,
]

\addplot[mark=o, mark options={scale=2.5}, blue]
  coordinates{
(100, 70.99)
(500, 73.11000000000001)
(1000, 76.95)
(2000, 79.59)
(3000, 80.34)
(4000, 80.04)
(5000, 80.33)
}; \addlegendentry{\huge Last}

\addplot[mark=o, mark options={scale=2.5}, red]
  coordinates{
(100, 84.16738372507513)
(500, 85.68586169342709)
(1000, 88.02532628485191)
(2000, 89.31440646663411)
(3000, 89.52459278504173)
(4000, 89.37344947345822)
(5000, 89.65973213994036)
}; \addlegendentry{\huge Average}

\end{axis}
\end{tikzpicture}{\label{subfig:sample-eff}}
    }}
    \subfloat[][\footnotesize ]{
    \resizebox{0.22\textwidth}{!}{
        \begin{tikzpicture}
\pgfplotsset{
    scale only axis,
    tick label style={font=\huge},
}

\begin{axis}[
  ymin=50, ymax=100,
  xmin=0, xmax=.5,
  xtick={0, .1, .2, .3, .4, .5},
    ymajorgrids=true,
    grid style=dashed,
  xlabel={\Huge Corruption Ratio},
  ylabel={\Huge Accuracy (\%)},
  legend pos=south east,
]

\addplot[mark=o, mark options={scale=2.5}, blue]
  coordinates{
  (0.0, 81.67999999999999)
(0.1, 76.69)
(0.2, 73.74000000000001)
(0.3, 72.16)
(0.4, 70.00999999999999)
(0.5, 69.01)
}; \addlegendentry{\huge Last}

\addplot[mark=o, mark options={scale=2.5}, red]
  coordinates{
  (0.0, 90.44186060115773)
(0.1, 87.82064506008955)
(0.2, 86.47664302773633)
(0.3, 85.55529300862561)
(0.4, 84.75042628897097)
(0.5, 83.40746815311289)
}; \addlegendentry{\huge Average}

\end{axis}
\end{tikzpicture}{\label{subfig:label-corr}}
    }}
    \caption{Performance of {\X} on Biased MNIST dataset (a) with varying number of samples per class; and (b) at different label corruption ratios.}
    \label{fig:label_corr}
\end{figure}

\noindent \textbf{Sample Efficiency}. In this experiment, we evaluate the sample efficiency of {\X} by controlling the number of instances per class. We perform this experiment on Biased MNIST dataset ($p=0.8$), where the number of instances per class is mostly uniform. In Figure~\ref{subfig:sample-eff}, we report the average and final accuracy achieved by {\X}. We observe that {\X} is quite sample efficient achieving similar performance with 1000 samples/class (originally there were $\sim6000$ samples/class). {\X} achieves strong performance even with as small as 100 samples/class.


\noindent \textbf{Label Corruption}. In this experiment, we evaluate the robustness of {\X} by corrupting labels in the training data. Specifically, we corrupted varying fractions of target attribute labels in the training set and report the average and final performance. In Figure~\ref{subfig:label-corr}, we observe a graceful degradation in performance as the label corruption ratio increases. Even at high corruption ratios, we observe that {\X} shows strong average performance, which showcases the robustness of our representation learning system. 

\begin{figure}[h!]
    \centering
    \resizebox{0.35\textwidth}{!}{
    \begin{tikzpicture}
\pgfplotsset{
    scale only axis,
    tick label style={font=\Large},
}

\begin{axis}[
  ymin=70, ymax=102,
  xmin=5, xmax=50,
  xtick={10, 20, 30, 40, 50},
  ytick={},
    ymajorgrids=true,
    grid style=dashed,
  xlabel={\huge \# Exemplars/Class},
  legend pos=outer north east,
  ylabel={\huge Accuracy (\%)},
]

\addplot[smooth, mark=o,mark options={scale=2}, red]
  coordinates{
(5, 89.89887815762532)
(10, 90.17622388255202)
(15, 89.90421954491174)
(20, 90.44)
(25, 89.83871829951805)
(30, 89.80336694124331)
(35, 89.85541351084987)
(40, 89.65479074654758)
(45, 89.98308588797626)
(50, 89.22127315679583)
}; \addlegendentry{\huge $p=0.80$}

\addplot[smooth, mark=o,mark options={scale=2}, blue]
  coordinates{
(5, 88.44597903740069)
(10, 88.22305181759025)
(15, 88.39342462340454)
(20, 88.21)
(25, 87.9445393264679)
(30, 87.8623320092252)
(35, 88.03869105033432)
(40, 87.58520528372787)
(45, 87.54120528372788)
(50, 87.10755771113504)
}; \addlegendentry{\huge $p=0.85$}

\addplot[smooth, mark=o,mark options={scale=2}, brown]
  coordinates{
(5, 84.57826864907307)
(10, 83.86434687512491)
(15, 83.86843775736756)
(20, 84.38)
(25, 83.62865591743096)
(30, 83.66328959479577)
(35, 83.66328959479577)
(40, 83.14979531794556)
(45, 83.21768444464566)
(50, 82.79026299249485)
}; \addlegendentry{\huge $p=0.90$}

\addplot[smooth, mark=o,mark options={scale=2}, black]
  coordinates{
(5, 76.54991144311539)
(10, 75.65866670196245)
(15, 76.01735605403141)
(20, 75.28)
(25, 75.66194070980345)
(30, 75.9970404453065)
(35, 75.21153689269633)
(40, 74.7505813928182)
(45, 75.44024471964389)
(50, 74.0820134123402)
}; \addlegendentry{\huge $p=0.95$}

\end{axis}
\end{tikzpicture}{\label{subfig:ex-acc}}
    }
    \caption{{Target accuracy with varying exemplar sizes.}}
    \label{fig:exemplar-size}
\end{figure}
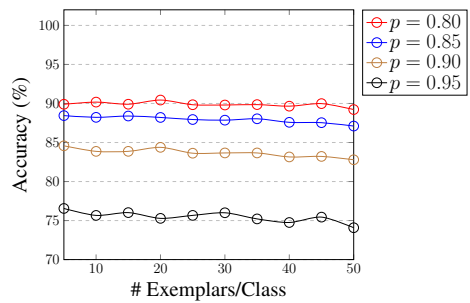

\noindent \textbf{Effect of exemplar size}. We perform ablation experiments on Biased MNIST to investigate the effect of exemplar set size on {\X}'s performance. In Figure~\ref{fig:exemplar-size}, we observe that the average target accuracy almost remains constant with different exemplar sizes in all settings. This shows that {\X} is robust to exemplar set size and can perform well with as small as $5$ samples/class. 

\section{Additional Results}

\begin{table*}[t!]
	\centering
	\resizebox{0.8\textwidth}{!}{
\begin{tabular}{@{}cl@{}| c c | c c|c c| c c}
    \toprule[1pt]
    	\multicolumn{2}{l@{~}|}{\multirow{2}{*}{\textbf{Method}}} & \multicolumn{2}{c|}{ $p=0.8$} & \multicolumn{2}{c|}{$p=0.85$} & \multicolumn{2}{c|}{$p=0.9$} & \multicolumn{2}{c}{$p=0.95$} \Tstrut\\
     & & DP & $\mathrm{Gap}_{\mathbf g}^{\mathrm{RMS}}$ & DP & $\mathrm{Gap}_{\mathbf g}^{\mathrm{RMS}}$ & DP & $\mathrm{Gap}_{\mathbf g}^{\mathrm{RMS}}$ & DP & $\mathrm{Gap}_{\mathbf g}^{\mathrm{RMS}}$ \Bstrut\\
    \midrule[0.5pt]
    \parbox[t]{0.8mm}{\multirow{7}{*}{\rotatebox[origin=c]{90}{\small{\texttt Incremental}}}} &  {LwF \citep{li2017learning} } & 0.19 & 0.046 & 0.29 & 0.124 & 0.56 & 0.230 & 1.57 & 0.548\\
    & Adversarial LwF & 0.36 & 0.148 &  0.15 & 0.098 & 1.01 & 0.398 & 0.77 & 0.298 \\
   &   {iCaRL \citep{rebuffi2017icarl}  } & 0.33 & 0.263 & 0.47 & 0.326 & 0.43 & 0.334 & \textbf{0.29} & \textbf{0.170} \Bstrut\\
     \cmidrule[0.5pt]{2-10}
    &  {\X} (\texttt{random}) & 0.13 & \textbf{0.047} & {0.12} & 0.067 & \textbf{0.18} & \textbf{0.087} & {0.50} & {0.237} \Tstrut\\
    &  {\X} (\texttt{prototype}) & {0.12} & 0.048 & 0.14 & \textbf{0.058} & 0.20 & 0.105 & 0.56 & 0.249\\
    &  {\X} (\texttt{submod})  & \textbf{0.12} & 0.050 & \textbf{0.12} & 0.070 & 0.22 & 0.114 & 0.63 & 0.305 \Bstrut\\
    \bottomrule[1pt]
\end{tabular}
}
	\caption{ {Amortized Demographic Parity and $\mathrm{Gap}_{\mathbf g}^{\mathrm{RMS}}$ scores of incremental and joint learning systems on Biased MNIST dataset. {\X} achieves the best performance among incremental learning baselines. Best results shown in \textbf{bold}.}}
	\label{tab:bmnist-dp}
\end{table*}
\noindent\textbf{Biased MNIST}.
\label{sec:addl-bmnist} We report the fairness metrics -- average DP and $\mathrm{Gap}_{\mathbf g}^{\mathrm{RMS}}$ scores for incremental learning systems on Biased MNIST dataset in Table~\ref{tab:bmnist-dp}. We observe that {\X} outperforms the baseline approaches in most settings except $p=0.95$. In the setting with $p=0.95$, iCaRL achieves the best score by a small margin which can be attributed to its relatively poor test accuracy.

\begin{figure}[h!]
	\centering
	\subfloat[][\footnotesize ]{
		\resizebox{0.20\textwidth}{!}{
			\begin{tikzpicture}
\pgfplotsset{
    scale only axis,
    tick label style={font=\huge},
}

\begin{axis}[
title={\Huge Leakage ($\downarrow$)},
  ymin=00, ymax=50,
  xmin=1, xmax=6,
  xtick={1, 2, 3, 4, 5, 6},
  ytick={10, 20, 30, 40},
    ymajorgrids=true,
    grid style=dashed,
  xlabel={\Huge Stage},
  legend pos=north east,
]

\addplot[mark=o, mark options={scale=2.5}, black]
  coordinates{
(1, 7.49514890744959)
(2, 10.071996682875337)
(3, 10.01310293080084)
(4, 9.738078545043585)
(5, 9.211331375428301)
(6, 9.503398753454235)
}; 

\addplot[mark=o, mark options={scale=2.5}, red]
  coordinates{
(1, 42.49894541466296)
(2, 45.241056956538124)
(3, 45.256627059231974)
(4, 44.382239988279984)
(5, 44.3424702235275)
(6, 44.370373358613875)
}; 

\addplot[mark=o, mark options={scale=2.5}, blue]
  coordinates{
(1, 27.620011811355788)
(2, 19.837410632389698)
(3, 21.660152531553422)
(4, 20.331090484811064)
(5, 17.9974302496329)
(6, 20.179720434879087)
}; 


\end{axis}
\end{tikzpicture}{\label{subfig:bios-leakage}}
	}}
	\subfloat[][\footnotesize]{
		\resizebox{0.28\textwidth}{!}{
			\begin{tikzpicture}
\pgfplotsset{
    scale only axis,
    tick label style={font=\huge},
    legend cell align={left},
}

\begin{axis}[
title={\Huge DP ($\downarrow$)},
  ymin=00, ymax=.8,
  xmin=1, xmax=6,
  xtick={1, 2, 3, 4, 5, 6},
  ytick={.15, .30, .45, .60, .75},
    ymajorgrids=true,
    grid style=dashed,
  xlabel={\Huge Stage},
  legend pos=outer north east,
]

\addplot[mark=o, mark options={scale=2.5}, black]
  coordinates{
(1, 0.13457754637477878)
(2, 0.3523615596264879)
(3, 0.3824499878003429)
(4, 0.3969707345362234)
(5, 0.4059212093164308)
(6, 0.41219041657142896)
}; \addlegendentry{\huge {\X}}

\addplot[smooth, mark=o, mark options={scale=2.5}, red]
  coordinates{
(1, 0.13464065768261582)
(2, 0.3345234243353054)
(3, 0.30692946530061277)
(4, 0.7163404917867086)
(5, 0.37005655918491587)
(6, 0.3141597558413642)
}; \addlegendentry{\huge Adv. LwF}

\addplot[mark=o, mark options={scale=2.5}, blue]
  coordinates{
(1, 0.13468539422781992)
(2, 0.3334694019530971)
(3, 0.3080619169933561)
(4, 0.39388746689199766)
(5, 0.39937372701135654)
(6, 0.25260101787486944)
}; \addlegendentry{\huge LwF}

\addplot[mark=o, mark options={scale=2.5}, brown]
  coordinates{
(1, 0.13468539422781992)
(2, 0.35869180697127373)
(3, 0.39159478504423983)
(4, 0.4313632427526617)
(5, 0.44634204883407413)
(6, 0.4484000164865532)
}; \addlegendentry{\huge iCaRL}

\end{axis}
\end{tikzpicture}{\label{subfig:bios-dp}}
	}}
	\caption{Evolution of gender leakage and Demographic parity across training stages using different incremental learning systems on Biographies dataset.}
	\label{fig:dp-evolution}
\end{figure}
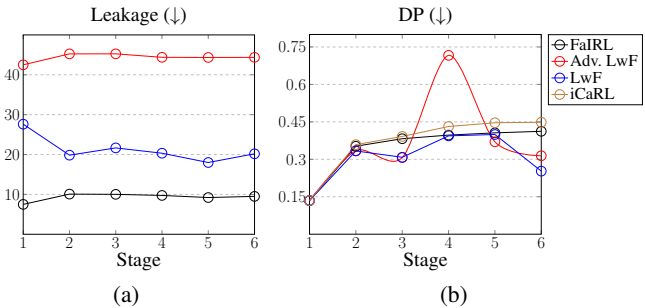

\noindent\textbf{Biography classification}.
\label{sec:bios-results} We report the gender leakage and demographic parity of incremental learning systems across different training stages in Figure~\ref{fig:dp-evolution} (expected trends for the metrics are shown using $\uparrow/\downarrow$). Gender leakage is evaluated by probing the representations for protected attribute. {We do not report the leakage for iCaRL as it performs prediction using nearest class estimator (NCE).} 
In Figure~\ref{subfig:bios-leakage}, we observe that {\X} leaks the least amount of information among the baselines. In Figure~\ref{subfig:bios-dp}, we observe that {\X} performs fairly well on demographic parity only falling short of LwF-based systems that suffer from catastrophic forgetting. {\X} consistently outperforms iCaRL, which achieves a good performance on classification task.

\section{Broader Impacts and Limitations}
\label{sec:impact}
Machine learning systems have revolutionized recommendations and decision-making in several applications like social media, over-the-top media industry etc. Over the years learning systems are found in critical applications like hiring, criminal recidivsm etc.  There are several positive benefits of having data-driven fair machine learning systems in such systems. However, a major drawback of such systems is that they are trained offline using prior data, tested in a single domain, and then deployed in the wild. Often it is expensive to monitor and control decisions made by such systems. In recent years, a large body of works have focused on developing systems that are fair towards different demographic groups. However, the efficacy of these approaches are unknown when faced with unseen data. There is a need for having systems that can continually adapt to new data in the wild. We believe that our proposed problem of developing incrementally trained fair machine learning systems is a first step towards this goal. Incrementally trained fair systems would pave the way for having systems that drive decisions in the wild, and enable organizations to deploy them in critical applications.

\noindent \textbf{Limitations}. We present {\X}, a novel fair representation learning system that can be trained in an incremental manner. One of the limitation of our system is that it relies on having protected attribute annotation to debias representations. Such annotations are expensive, and may not be accurate in all scenarios. For example, in our experiments we use only binary gender labels without considering other possible gender identities. Prior works have also shown that debiasing a system w.r.t a single protected attribute can sometimes make it biased towards other demographic groups. {\X} relying on protected labels can also hinder learning of the target task, when the protected and  target attributes are strongly correlated. Therefore, it is important for future systems to identify biased shortcuts used by a model, and prevent the system from using them for final predictions.

\noindent\textbf{Negative Usage}. We proposed {\X} with the intention of achieving fairness using machine learning systems in the real-world. However, our approach can be misused if developers it to learn sensitive information instead of debiasing representations from it. In such cases, it easy to flag such systems using the fairness metrics discussed in Section~\ref{sec:metrics}.

We hope that our proposed task of incrementally learning fair systems would encourage others to develop more robust systems and help achieve fairness in the wild.  



\end{document}